\documentclass[acmtog]{acmart}

\usepackage{booktabs} 
\usepackage{enumitem}
\usepackage{amsmath}
\usepackage{multirow}
\usepackage{xcolor}
\usepackage{float}
\usepackage{balance}

\DeclareMathOperator*{\argmin}{arg\,min}

\definecolor{Tomato}{RGB}{255, 99, 71}
\definecolor{DodgerBlue}{RGB}{30, 144, 255}
\newcommand{\OursNet}{InstantDrag}

\usepackage[ruled]{algorithm2e} 

\SetAlFnt{\small}
\SetAlCapFnt{\small}
\SetAlCapNameFnt{\small}
\SetAlCapHSkip{0pt}

\copyrightyear{2024}
\acmYear{2024}
\setcopyright{rightsretained} 
\acmConference[SA Conference Papers '24]{SIGGRAPH Asia 2024 Conference Papers}{December 3--6, 2024}{Tokyo, Japan} 
\acmBooktitle{SIGGRAPH Asia 2024 Conference Papers (SA Conference Papers '24), December 3--6, 2024, Tokyo, Japan}
\acmDOI{10.1145/3680528.3687668} 
\acmISBN{979-8-4007-1131-2/24/12}

\begin{document}
\title{InstantDrag: Improving Interactivity in Drag-based Image Editing}

\author{Joonghyuk Shin}
\orcid{0000-0003-3780-3897}
\affiliation{
 \institution{Seoul National University}
 \country{South Korea}}
\email{joonghyuk@snu.ac.kr}

\author{Daehyeon Choi}
\orcid{0009-0003-1438-4949}
\affiliation{
 \institution{POSTECH}
 \country{South Korea}}
\email{daehyeonchoi@postech.ac.kr}

\author{Jaesik Park}
\orcid{0000-0001-5541-409X}
\affiliation{
 \institution{Seoul National University}
 \country{South Korea}}
\email{jaesik.park@snu.ac.kr}




\begin{abstract}
\label{sec:0}
Drag-based image editing has recently gained popularity for its interactivity and precision. 
However, despite the ability of text-to-image models to generate samples within a second, drag editing still lags behind due to the challenge of accurately reflecting user interaction while maintaining image content.
Some existing approaches rely on computationally intensive per-image optimization or intricate guidance-based methods, requiring additional inputs such as masks for movable regions and text prompts, thereby compromising the interactivity of the editing process.
We introduce \OursNet, an optimization-free pipeline that enhances interactivity and speed, requiring only an image and a drag instruction as input.
\OursNet~consists of two carefully designed networks: a drag-conditioned optical flow generator (FlowGen) and an optical flow-conditioned diffusion model (FlowDiffusion). \OursNet~learns motion dynamics for drag-based image editing in real-world video datasets by decomposing the task into motion generation and motion-conditioned image generation. We demonstrate \OursNet's capability to perform fast, photo-realistic edits without masks or text prompts through experiments on facial video datasets and general scenes. 
These results highlight the efficiency of our approach in handling drag-based image editing, making it a promising solution for interactive, real-time applications.
\end{abstract}


\begin{teaserfigure}
    \centering
    \includegraphics[width=\textwidth]{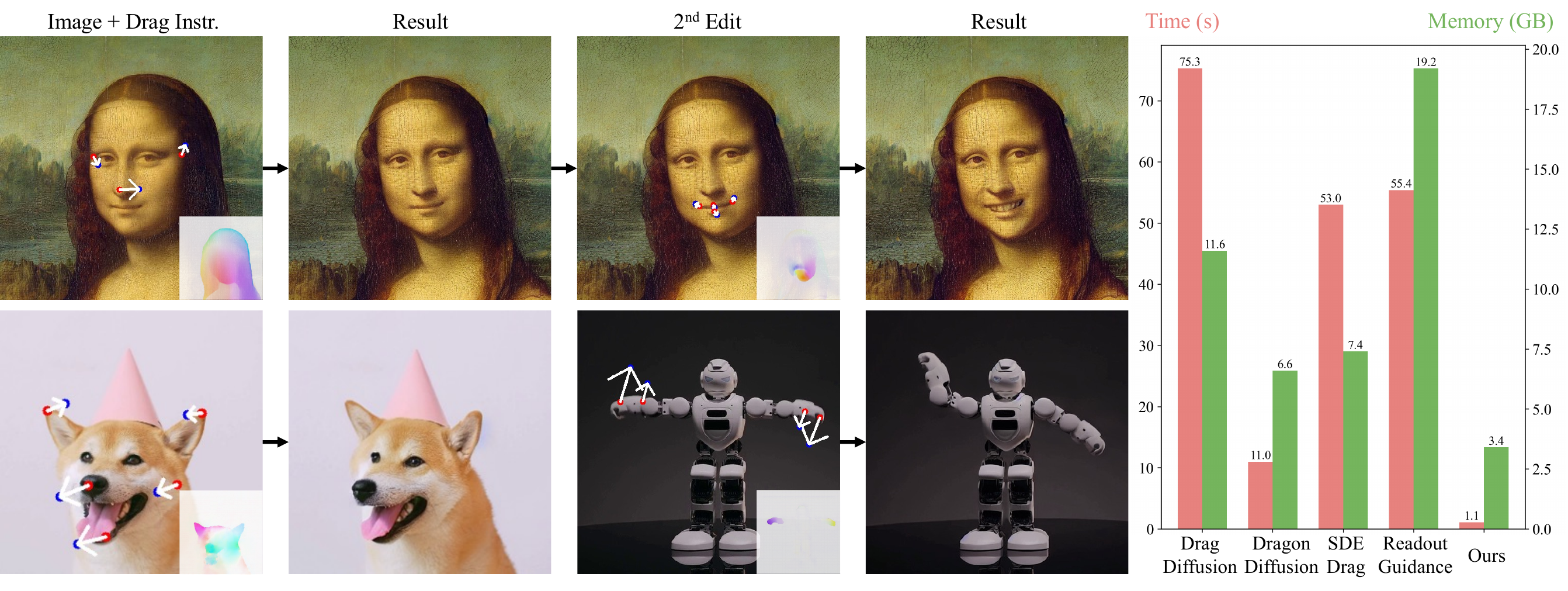}
    \caption{We introduce InstantDrag, an efficient framework that combines an optical flow generator with a motion-conditioned diffusion model. Our approach enables realistic drag edits in roughly a second, leveraging real-world video datasets. (Inputs: \textcircled{c} Louvre Museum\textsuperscript{\ref{fn:louvre}}, \textcircled{c} A. Shvets and P. Danilyuk, via Pexels)}
    \vspace{2mm}
    \label{fig:teaser}
\end{teaserfigure}

\maketitle
\footnotetext[1]{\label{fn:louvre}\textcircled{c} 2011 GrandPalaisRmn  (musée du Louvre) / Michel Urtado. Mona Lisa. Louvre Collections: https://collections.louvre.fr/ark:/53355/cl010062370}
\section{Introduction}
\label{sec:1}

Recent advancements in diffusion models~\cite{ho2020denoising, song2021scorebased} have empowered users to transform their creative visions into digital images using only a few text prompts. Further research has proposed faster ODE solvers~\cite{lu2022dpm}, consistency distillation~\cite{song2023consistency}, straightening of ODE trajectories~\cite{liu2022flow}, and adversarial learning~\cite{sauer2023adversarial} to improve inference speed. However, while text-to-image generation can now be achieved in a fraction of a second, editing real images with precision and interactivity remains relatively slow and less performant.

Text-guided editing techniques allow users to easily modify high-frequency features and the overall mood of an image, but accurately pinpointing the region of interest for editing remains challenging. Drag editing, introduced in the pioneering work DragGAN~\cite{pan2023drag}, directly operates in pixel space, granting users more precise control and the ability to edit low-frequency features, including motion-aware structural changes. This increased editing capability comes at the cost of increased complexity. Specifically, drag-based editing requires the model to have a comprehensive understanding of potential motion information, posing different challenges compared to text-based modifications. Consequently, state-of-the-art approaches usually start from LoRA training~\cite{hu2021lora}, DDIM inversion~\cite{song2020denoising} and perform computationally intensive latent optimization or intricate score guiding to capture and manipulate motion dynamics.

Despite the core motivation of drag-based image editing being the interactivity of the editing process, we observe that most existing works are based on complex optimization techniques, which can fundamentally slow down the editing speed and hinder interactivity. To address this, we propose a new approach that focuses on four key aspects: 1) speed, 2) real image editing quality, 3) removing the reliance on user input masks to define movable regions, and 4) eliminating the necessity of text prompts.

To handle these challenges, we present \textit{\OursNet}, a pipeline designed to enhance interactivity and speed without optimization, taking only an image and a drag instruction as input. We decouple the drag-editing task into motion generation and motion-conditioned image generation, assigning each task to a generative model with the appropriate capacity and design. FlowGen, a lightweight and fast GAN-based model, generates dense optical flow from sparse user inputs. FlowDiffusion, an efficiently tuned optimization-free diffusion model, performs high-quality edits conditioned on the generated motion. Our method achieves high-quality results in roughly a second by streamlining the architecture and removing redundant components such as the text encoder in the diffusion model.

One significant challenge in this setting is the lack of fine-grained paired datasets of drag instructions and edited images. The na\"ive use of video datasets can result in the undesired movements of non-target objects, including the background, reducing the precision. We carefully analyze and ablate motion cues in videos and experimentally demonstrate the effectiveness of our training strategy in learning motion dynamics for drag-based image editing.

We conduct comprehensive experiments to validate the strength of each module in InstantDrag. Our experiments on facial video datasets and general scenes demonstrate that our method can deliver photo-realistic edits on real images without requiring metadata, such as user input masks or text prompts. Compared to recent works, editing real images with InstantDrag is up to \(\sim\)\textit{75\(\times\)} faster while consuming up to \(\sim\)\textit{5\(\times\)} less GPU memory. Fig.~\ref{fig:teaser} shows our results and compares computational complexity with other approaches.

\section{Related Work}
\label{sec:2}

\subsection{Image Editing with Generative Models}
\label{sec:2.1}
Generative adversarial networks (GANs) have achieved significant successes in image generation~\cite{goodfellow2014generative}, laying the foundation for several seminal works in the field of image editing. One prominent approach, image-to-image translation, focuses on mapping an image from a source domain to a target domain. Numerous methods have been proposed for both paired settings~\cite{Isola_2017_CVPR} and unpaired settings~\cite{Zhu_2017_ICCV}. These approaches typically employ U-shaped networks to directly transform images with the aid of a discriminator to enhance finer details. A combination of reconstruction and adversarial losses guides the transformation process. Other approaches involve inverting an input image back to a continuous latent space of a pre-trained GAN and perform editing guided by informative features from external sources. However, editing general images with more complex instructions, such as text cues, remains challenging due to the inherent difficulties in scaling up GANs. Consequently, the generalizability of these methods is constrained by the representative power of GANs.

Rapid developments in large-scale text-to-image diffusion models~\cite{rombach2022high, ramesh2022hierarchical} have significantly improved image quality for general scenes by training on web-scale datasets~\cite{schuhmann2022laion}. Building upon these diffusion models, several approaches have been proposed to enable text-guided image editing~\cite{meng2021sdedit, hertz2022prompt, brooks2023instructpix2pix, parmar2024one, mokady2023null}. Recent advancements have extended these capabilities to 3D and movement-aware edits by leveraging 3D-lifted activations and learning from dynamic video data~\cite{pandey2024diffusion, alzayer2024magic}. Compared to GAN-based methods, diffusion-based generation and editing offers better stability and generalizability, albeit at the cost of increased computational load and memory consumption. 

\subsection{Drag-based Image Editing}
\label{sec:2.2}

DragGAN~\cite{pan2023drag} introduced an interactive image editing technique where users perform motion-aware edits by dragging source points to target points. This method optimizes the \(\mathcal{W^+}\) latent space of StyleGAN2~\cite{Karras_2020_CVPR} using iterative motion supervision and point tracking. Given source points \(\{s_i = (x_{s,i}, y_{s,i})\}\) and target points \(\{t_i = (x_{t,i}, y_{t,i})\}\), motion supervision moves a small patch around \(s_i\) to \(t_i\). The motion supervision loss is:
\begin{align}
    \mathcal{L} = \sum_{i=0}^n \sum_{q_i \in \Omega_1(s_i, r_1)} \left\| F(q_i) - F(q_i + d_i) \right\|_1 + \lambda \left\| (F - F_0) \cdot (1 - M) \right\|_1.
\end{align}
The first term drives pixel features around \(s_i\) to move towards \(t_i\) along \(d_i\), a normalized vector directing from source to target, and the second term maintains the unmasked features close to the initial features \(F_0\). Features \(F(x)\) are from StyleGAN2's 6th block. After back-propagating to update the latent code \(w'\), point tracking finds the new source point \(s_i'\) via nearest neighbor search:
\begin{align}
    s_i' := \argmin_{q_i \in \Omega_2(s_i, r_2)} \| F(q_i) - F_0(s_i) \|.
\end{align}

Several works have extended this iterative latent optimization framework using motion supervision and point tracking. DragDiffusion~\cite{shi2023dragdiffusion} adapts DragGAN to text-to-image diffusion models, enabling editing of general scenes. FreeDrag~\cite{ling2023freedrag} addresses ambiguous point tracking by incorporating adaptive template features and line search with backtracking. More recent approaches such as GoodDrag~\cite{zhang2024gooddrag}, Drag Your Noise~\cite{liu2024drag}, StableDrag~\cite{cui2024stabledrag}, and EasyDrag~\cite{hou2024easydrag} enhance the DragDiffusion framework by alternating drag and denoising operations, leveraging diffusion semantic propagation, introducing discriminative point tracking methods with confidence-based motion supervision, and applying more stable motion supervision. Other approaches focus on guidance from external networks~\cite{mou2023dragondiffusion, luo2024readoutguidance}, or copy \& paste style latent manipulation with SDE formulation~\cite{nie2023blessing}.

While these approaches demonstrate impressive quality edits, they have limitations in terms of interactivity. A typical latent optimization-based drag editing framework, like DragDiffusion, comprises 1) LoRA training of U-Net for enhanced identity preservation, 2) DDIM inversion for real images, 3) latent optimization using motion supervision and point tracking, and 4) diffusion sampling process. These steps can take tens of seconds to minutes, depending on the drag instruction and image size. Similarly, guidance-based methods like DragonDiffusion and Readout Guidance also encounter computational bottlenecks. These methods require DDIM inversion followed by a long sampling process, guided by either an additional diffusion branch or backpropagated gradients from external networks. For some works that involve manual drawing of movable region masks and prompt engineering, the actual editing process can be even longer. Furthermore, during the DDIM inversion process, high-frequency details are not accurately inverted, resulting in notable performance degradation with real images.

To address these limitations, we aim to build a more interactive pipeline where images can be dragged without optimization and metadata by training dedicated models. Concurrent work~\cite{li2024dragapart} adopts a similar optimization-free editing idea by training on synthetic objects, but it primarily focuses on part-level movements of relatively simple objects with white backgrounds. In contrast, our work focuses on dragging real-world images with complex backgrounds. Our motion-conditioned diffusion model shares similarities with motion guidance~\cite{geng2023motion} in its utilization of optical flow. However, their training-free method uses a differentiable off-the-shelf optical flow network to guide the sampling process. In contrast, our method primarily trains a dedicated diffusion model to accept flow conditions and perform classifier-free guidance~\cite{ho2021classifier}, resulting in faster sampling.

\begin{figure}[t]
    \includegraphics[width=\linewidth]{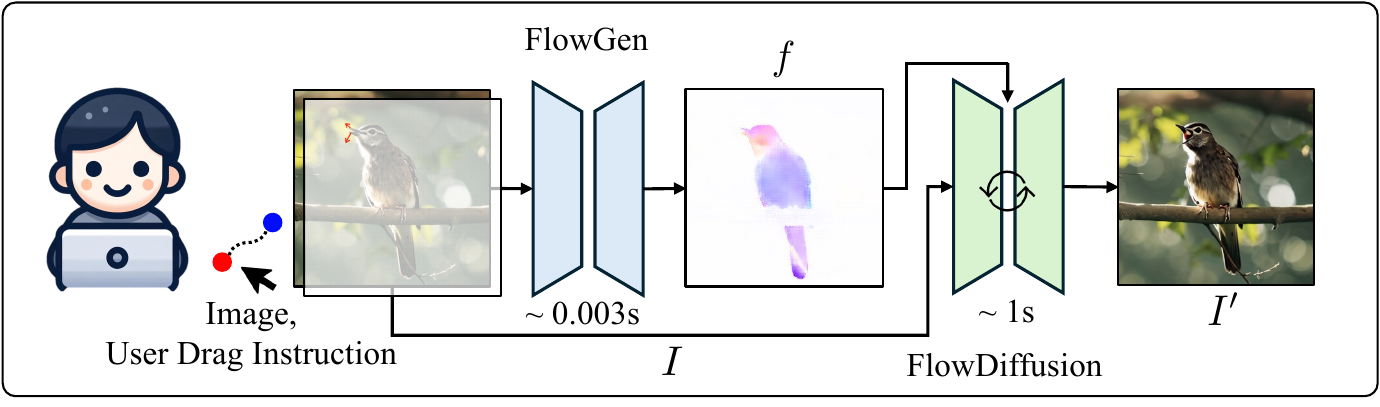}
    \caption{Illustration of our inference pipeline. Given sparse user drag input, our FlowGen estimates dense optical flow, and our FlowDiffusion edits the original image with the flow guidance. Our approach does not require auxiliary input, such as texts or foreground masks. Our approach is inversion and optimization-free, providing the edited image in about a second.}
    \vspace{-1mm}
    \label{fig:Inf_pipeline}
\end{figure}

\section{Method}
\label{sec:3}

\subsection{Model Architectures}
\label{sec:3.1}

We tackle drag-editing by dividing the task into two components: motion generation and motion-conditioned image generation. Each component is handled by a specialized network: a GAN-based network, FlowGen, for motion generation, and a diffusion-based network, FlowDiffusion, for motion-conditioned image generation. The overview of our inference pipeline is illustrated in Fig.~\ref{fig:Inf_pipeline}.

\subsubsection{FlowGen}
\label{sec:3.1.1}
\begin{figure}[t]
    \centering
    \includegraphics[width=\linewidth]{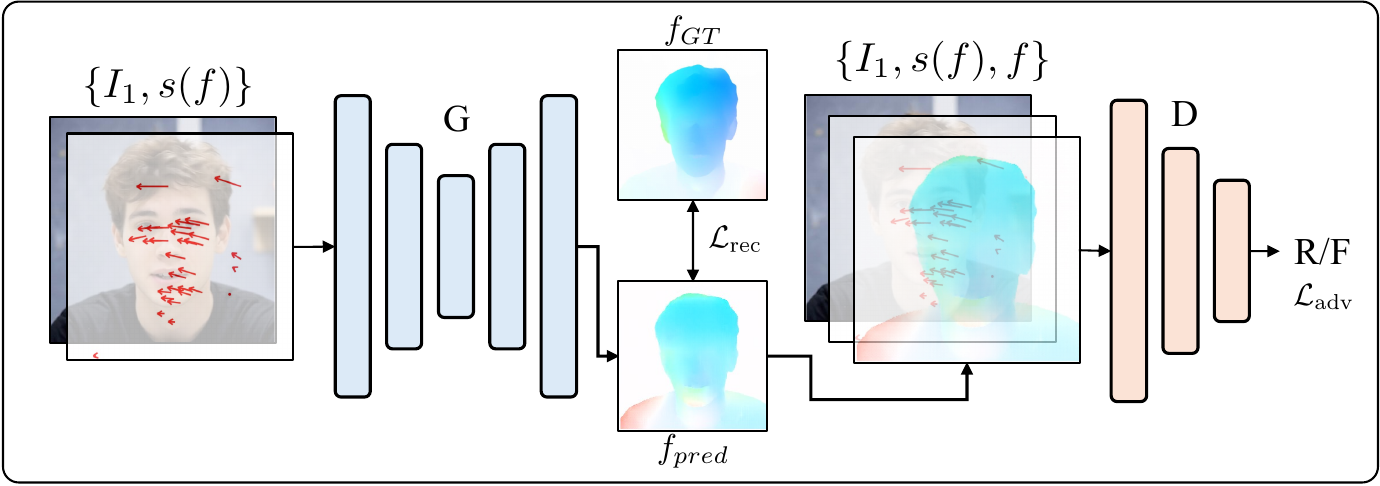}
    \caption{Illustration of our FlowGen architecture (Sec.~\ref{sec:3.1.1}). Sparse user drag input is channel-wise concatenated with the input image and fed into the generator, which predicts dense optical flow. Based on a Pix2Pix-like GAN architecture, FlowGen is trained using the adversarial loss from the discriminator and the reconstruction loss from the generator.}
    \label{fig:FlowGAN}
    \vspace{-2mm}
\end{figure}

Our rationale for employing a GAN in motion generation is based on conceptualizing this task as a translation problem, where the goal is to map an RGB image with drag instructions (sparse flow) to a dense optical flow. Inspired by the success of GAN-based models like Pix2Pix~\cite{Isola_2017_CVPR}, we recognize that this translation task can be efficiently performed with a one-step generative model, which also helps reduce the editing time. Consequently, we opted to train our model from scratch in a Pix2Pix-style manner. As illustrated in Fig.~\ref{fig:FlowGAN}, our generator receives a total of 5 channels of input data: 3 channels for the input image and 2 channels for the condition sparse drag instructions. It outputs a 2-channel dense optical flow. On the other hand, the discriminator processes a 7-channel input comprising 3 channels of input image, 2 channels of sparse flow, and 2 channels of dense flow.

To accelerate training and increase the model capacity for larger datasets, we employ GroupNorm instead of InstanceNorm and utilize a deeper architecture for the PatchGAN-based discriminator. To encourage the generator to learn a robust mapping from various sparse flows to dense flow, we update the generator four times using randomly sampled sparse flows for each update of the discriminator. More details can be found in the Appendix. Given the initial frame \( x \), dense optical flow \( f \), and a sparse flow \( f_s \) sampled using the sampler \( S \) (i.e., \( f_s = S(f) \)), FlowGen's adversarial $\mathcal{L}_{adv}(G,D)$ and reconstruction $\mathcal{L}_{rec}(G)$ losses can be written as follows:
\begin{align*}
    \mathcal{L}_{adv} = \mathbb{E}_{x,f} \left[ \log D(x, f_s, f) \right] + \mathbb{E}_{x,f} \left[ \log \left( 1 - D(x, f_s, G(x, f_s)) \right) \right],
\end{align*}
\begin{align}
    \mathcal{L}_{rec} = \mathbb{E}_{x,f} \left[ \left\| f - G(x, f_s) \right\|_2 \right].
\end{align}

\subsubsection{FlowDiffusion}
\label{sec:3.1.2}
\begin{figure}[t]
    \includegraphics[width=0.8\linewidth]{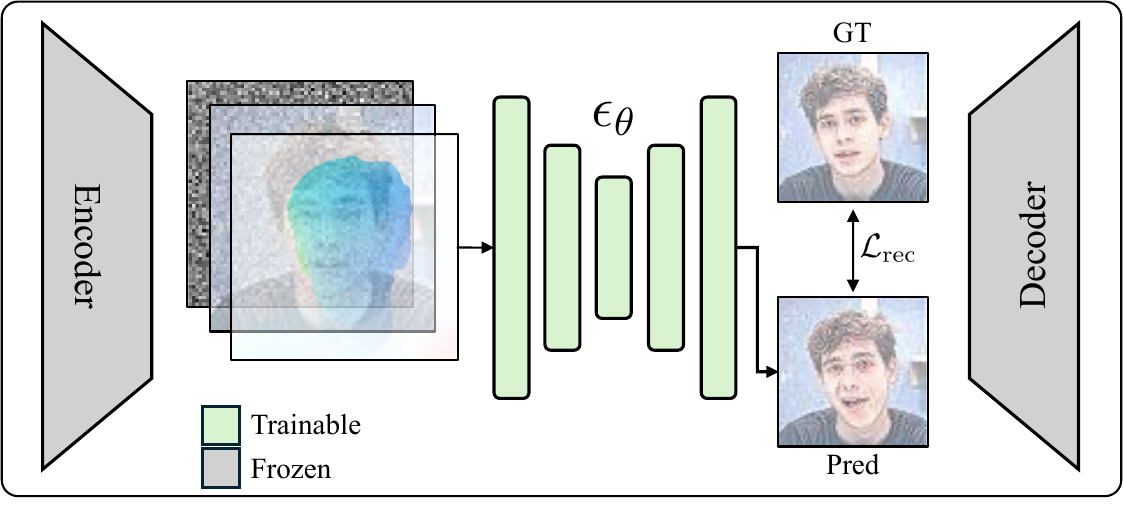}
    \caption{Illustration of our FlowDiffusion architecture (Sec.~\ref{sec:3.1.2}). The denoising U-Net of FlowDiffusion takes encoded image and downscaled optical flow as inputs. It leverages channel-wise concatenated input image and optical flow to guide the denoising process, learning to predict subsequent video frames in the latent space based on motion information.}
    \label{fig:FlowDiffusion}
    \vspace{-2mm}
\end{figure}

Making changes to reflect the motion condition is known to be challenging and time-consuming. Specifically, directly adopting off-the-shelf optical flow networks to guide the denoising process can take dozens of minutes~\cite{geng2023motion}. Therefore, we decide to train a diffusion model specifically designed to accept dense optical flow as a condition. Our baseline is set by Instruct-Pix2Pix~\cite{brooks2023instructpix2pix}, which incorporates conditioning of the input image and text prompts in fine-tuning Stable Diffusion. As detailed in Sec.~\ref{sec:3.2.4}, we discovered that simply concatenating the optical flow channels with the input image channels is highly effective, provided the optical flow is properly normalized. As shown in Fig.~\ref{fig:FlowDiffusion}, our U-net now has a 10-channel input, consisting of 4-channel latent noise, 4-channel latent image, and 2 channels for optical flow. 

A key difference in our approach compared to Instruct-Pix2Pix is that our edit signal is encoded in additional channels for the flow dimension, not in the text prompt. While Instruct-Pix2Pix needs to reflect signals from both text and image domains, our model needs to maintain consistency except for the dragged regions, reflecting only the dense flows without relying on textual input. 

We experimented with different guidance and training mechanisms, finding out that replacing text tokens with null tokens works reasonably well. This approach also has the advantage of reducing the computational costs associated with text encoders. Given our denoising network \(\epsilon_\theta\), a noisy latent \(z_t\), encoded input image \(c_I\), optical flow \(c_F\), and guidance scales \((s_I, s_F)\) for each conditional input, FlowDiffusion's classifier-free guidance with input image and optical flow can be written as follows:
\begin{align}
    \tilde{\epsilon}_\theta(z_t, c_I, c_F) = & \ \epsilon_\theta(z_t, \varnothing, \varnothing) \nonumber \\
    & + s_I \cdot [\epsilon_\theta(z_t, c_I, \varnothing) - \epsilon_\theta(z_t, \varnothing, \varnothing)] \nonumber \\
    & + s_F \cdot [\epsilon_\theta(z_t, c_I, c_F) - \epsilon_\theta(z_t, c_I, \varnothing)].
\end{align}
Unlike the image and text prompt relationship in the InstructPix2Pix pipeline, where conditional dropout occurs for 10\% in each case, with a 5\% overlap for dropping both, dropping the image only while using flow as input does not intuitively make sense. Therefore, for training of FlowDiffusion, we drop the image condition 5\% of the time and the flow condition 10\% of the time, ensuring the model is not conditioned solely on flow without the image.

\subsection{Implementation Details}
\label{sec:3.2}

\subsubsection{Preparing Datasets}
\label{sec:3.2.1}
The main challenge in training a drag-dedicated model, (\textit{i.e.} \(output = model(input, c_{drag})\)), is the lack of a curated dataset composed of triplets: input image, output image, and drag condition. Due to the difficulty of obtaining such edit pairs in image domains, we resort to video datasets. 

From each video, we extract frames and randomly apply a sliding window technique to sample pairs. Choosing an appropriate window size is crucial, as it determines the extent of motion that the model can learn; hence, it must be large enough to encapsulate realistic motions. For the large-scale facial video dataset CelebV-Text~\cite{yu2023celebv}, we sample frames at 10fps and use a maximum interval of 8 for sampling two pairs. Optical flows between these pairs are extracted using FlowFormer~\cite{huang2022flowformer}, and segmentation models are employed to obtain masks of the objects of interest. 

To demonstrate that our method is not limited to facial editing, we also pre-train our model on the widely used optical flow datasets FlyingChairs~\cite{dosovitskiy2015flownet} and SINTEL~\cite{butler2012naturalistic}. For general scenes, due to the computational constraints on training on full web-scale video datasets, we propose a novel setting where a short video (10\(\sim\)60s) is provided for model fine-tuning at test time. The resulting dataset is composed of $n$ pairs, each consisting of two images, two masks, and an optical flow directing frame 1 to 2: \(\{I_{1,i}, I_{2,i}, M_{1,i}, M_{2,i}, f_i\}_{i=1}^n\). For training FlowGen and FlowDiffusion, we use masked optical flow \(f_i \cdot M_{1,i}\) as GT to capture the object's motion exclusively.

\subsubsection{Pseudo Drag Instructions for FlowGen}
\label{sec:3.2.2}
\begin{figure}[t]
    \includegraphics[width=\linewidth]{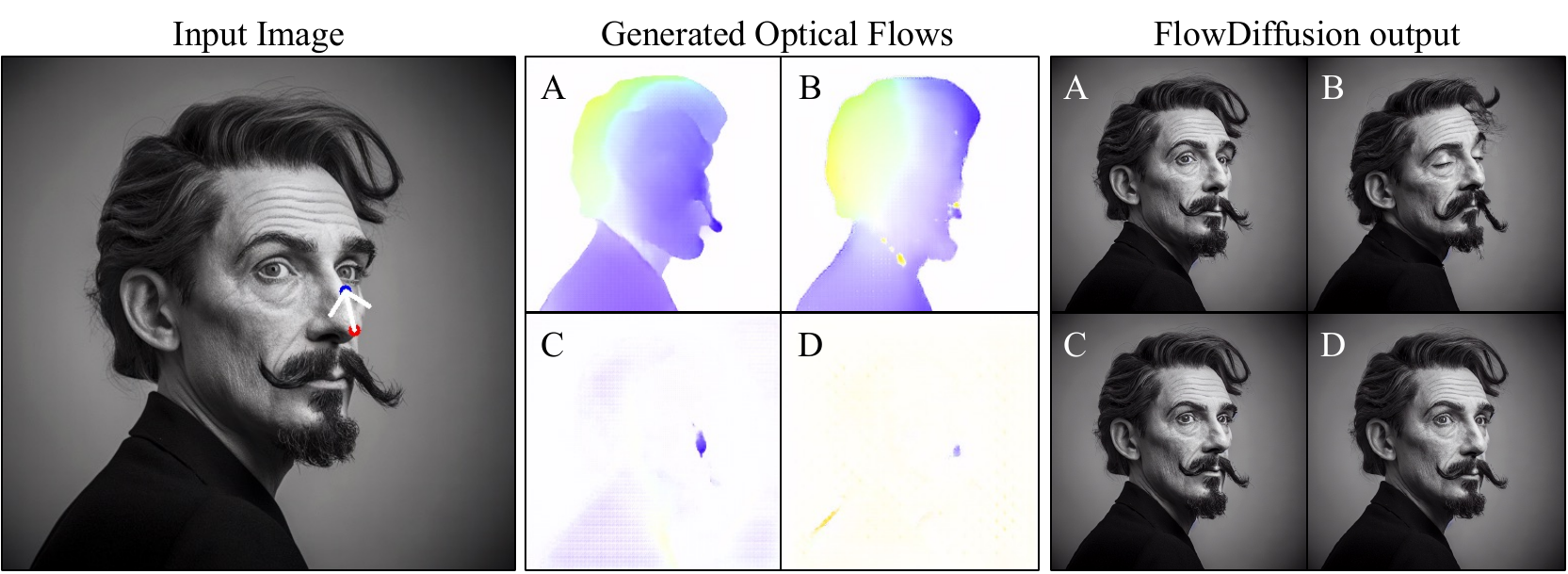}
    \caption{Dragging results from FlowGen trained under four settings: (A) Stochastic sampling strategy (Sec.~\ref{sec:3.2.2}), (B) 1 fixed point (nose), (C) 100 fixed grid points, (D) 900 fixed grid points. Excessive points (C, D) generate sparse motion while a single point (B) causes undesired movements. We find (A) to be the most robust, combining the advantages of the other approaches.}
    \label{fig:sparse_flow}
    \vspace{-2.5mm}
\end{figure}

Drag-based manipulation is an inherently ill-posed problem where there can be multiple valid movements for a given input. User inputs can range from a single point to an extensive set of points. We expect our model to be capable of handling this wide variation to generate plausible motion. Providing a single sparse point to encapsulate overall movements would require the generator to infer the entire motion across the image. For example, moving the nose might necessitate adjustments in other parts, like the eyes, leading to the stitching of uncorrelated movements. On the other hand, providing thousands of dense points, where each point captures local motion, would result in very localized movements. Achieving accurate results at test time with this approach would require an impractically large number of points (see Fig.~\ref{fig:sparse_flow}). Additionally, we need to sample points from areas of interest to ensure meaningful motion generation. Sampling points outside the target object would be ineffective.

To address these challenges, we conducted an extensive search on sampling strategies. We first initialize the sparse flow \(f_s \in \mathbb{R}^{2\times h\times w}\), with random values sampled from \(U(0,1)\). Then, we assign \(-\infty\) values to masked regions (background) while adding a value \(v_g\) to grid locations, and \(v_s\) to specific points such as facial keypoints. Finally, we select top-$k$ points in the sparse flow and zero out the remaining parts, where \(k\) is also randomly chosen in predefined range. It is important to note that the grid size plays a crucial role in preventing points from being selected only in certain areas. We use a subset of dlib keypoints as special points for facial videos and apply grid-based sampling exclusively for fine-tuning general scenes. To provide flexibility, we offer different GAN configurations depending on the user's intentions. For example, users can choose configurations for single-point dragging, keypoint-based fine-grained motion editing, or extremely fine-grained editing such as hair. This plug-and-play flexibility of FlowGen allows our model to handle varying levels of detail in user inputs, ensuring robust motion generation. Further details are available in the Appendix.

\subsubsection{Enforcing Background Consistency for FlowDiffusion}
\label{sec:3.2.3}
Another difficulty in directly adopting natural videos is maintaining background consistency. Unlike video generation, where any plausible movements are acceptable, drag-based image editing requires a consistent background and allows movements only in the dragged object. By using the binary mask obtained in Sec.\ref{sec:3.2.1}, we can enforce background consistency with mask-based operations.

Consider two frames, $I_1=I_{fg}$ and $I_2=I_{bg}$, with corresponding masks $M_1=M_{fg}$ and $M_2=M_{bg}$. We aim to create a composite image combining the object from $I_{fg}$ with the background from $I_{bg}$. To account for imperfect masks, we dilate $M_{bg}$ using a 15x15 kernel to expand the coverage of $I_{fg}$ in boundary regions. We define inverted masks $M_{fg}^{\text{inv}} = 1 - M_{fg}$ and $M_{bg}^{\text{inv}} = 1 - M_{bg}^{\text{dilated}}$ to hold 1 for background and 0 for the object. Our mask operation is as follows:
\begin{align}
I_{fg}^{\text{new}} &= I_{fg} \cdot M_{fg} \\
I_{bg}^{\text{new}} &= (M_{fg}^{\text{inv}} \land M_{bg}^{\text{inv}}) \cdot I_{bg} + [1 - (M_{fg} \lor M_{bg}^{\text{inv}})] \cdot I_{fg} \\
I^{\text{new}} &= I_{fg}^{\text{new}} + I_{bg}^{\text{new}}
\end{align}
\begin{figure}[t]
    \includegraphics[width=\linewidth]{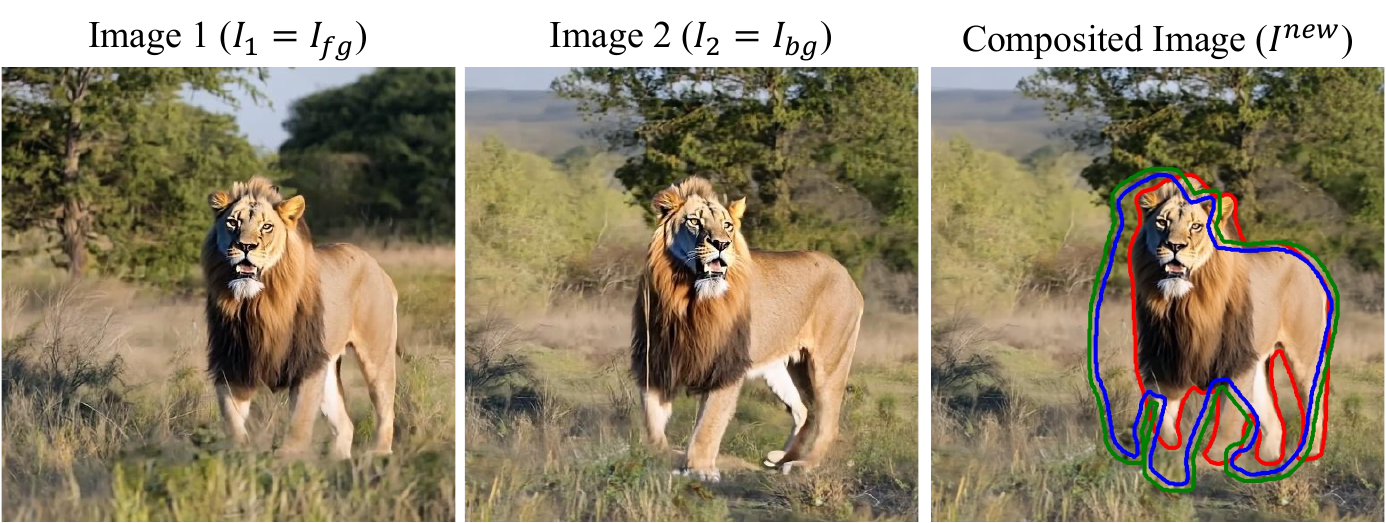}
    \caption{Visualization of the mask operation in Sec~\ref{sec:3.2.3}. $I^{new}$ combines the object from $I_{1}$ and the background from $I_{2}$. Blue contour shows $I_2$'s mask and green contour shows its dilated mask.}
    \label{fig:BG_consistency}
\end{figure}

The visualization of our method can be found in Fig.~\ref{fig:BG_consistency}. We explored three approaches: (1) compositing $I_2$'s background with $I_1$'s object, (2) compositing $I_1$'s background with $I_2$'s object, and (3) no background consistency. Ablation studies showed the first approach performed best. The intuition is as follows: when the diffusion model learns to transform $I_1$ to $I_2$ using flow $f$, it uses $I_1$ and $f$ as conditions and calculates the loss between its output and $I_2$. Although mask operations are generally robust, modifying the ground truth $I_2$ could lead to training with perturbed targets, potentially causing artifacts at test time. In contrast, modifying the input condition $I_1$ allows the model to handle and ignore these perturbations during the denoising process, resulting in a proper reconstruction of $I_2$. For this reason, we set $I_1$ as the foreground image and $I_2$ as the background image. Note that masks are only used during training, not inference.

\subsubsection{Optical Flow Normalization}
\label{sec:3.2.4}
\begin{figure}[t]
    \includegraphics[width=\linewidth]{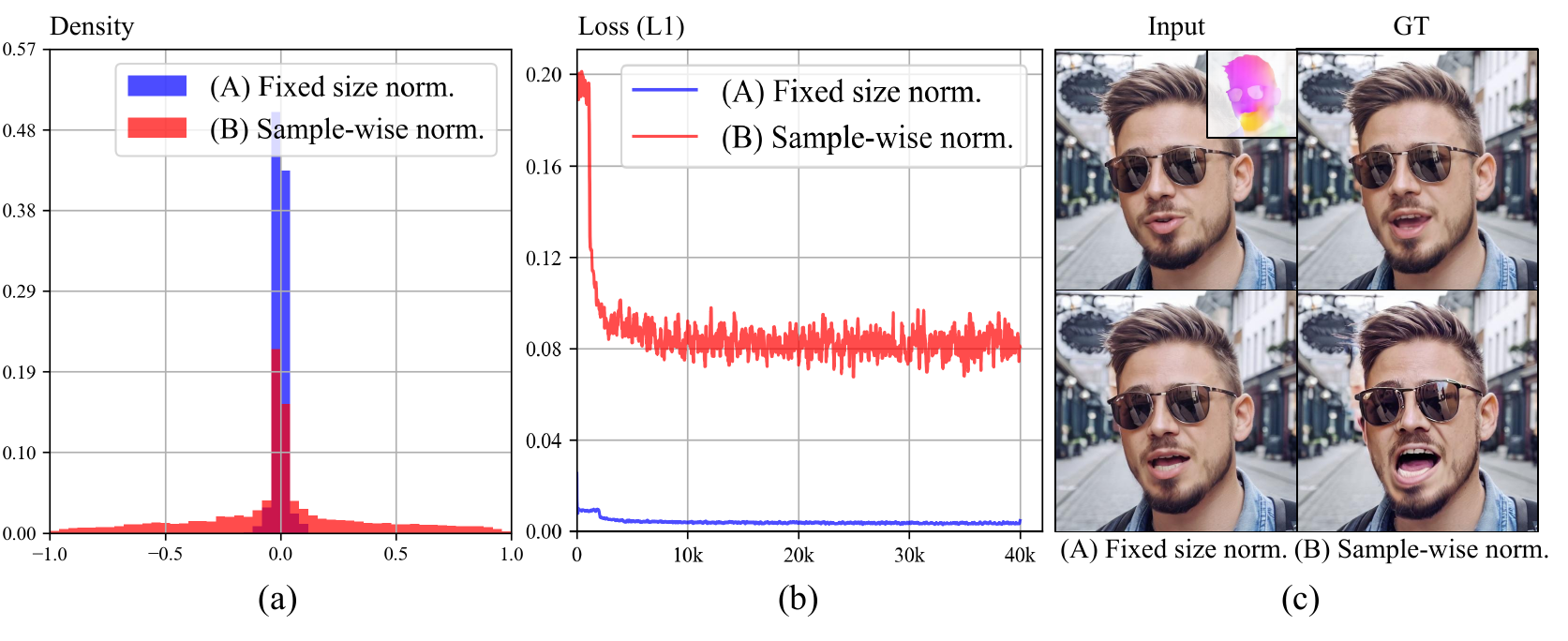}
    \caption{(a) Flow distribution after normalization. (b) L1 Reconstruction loss of FlowGen for two normalization methods. (c) Inference results of FlowDiffusion, showing that sample-wise normalization (B) does not consider the actual flow scale. Details are described in Sec.~\ref{sec:3.2.4}.}
    \label{fig:flow_normalization}
    \vspace{-5mm}
\end{figure}

Conditioning on optical flow directly without a warping operation is a relatively unexplored area. An optical flow \(f \in \mathbb{R}^{2\times h\times w}\) can range from \((-h, h)\) or \((-w, w)\). Directly calculating a loss or conditioning on this scale can lead to numerical errors due to the sensitive scales of model weights. Additionally, resizing the flow to the latent space causes scale inconsistencies. Therefore we consider two variants: (1) fixed size normalization where we divide \(f\) by the spatial dimensions and (2) sample-wise normalization where we divide each channels by its absolute maximum value. An advantage of fixed size normalization is that it preserves the actual size and scale, making the numbers directly proportional to the dimensions. However, as shown in Fig.~\ref{fig:flow_normalization}, this results in a very narrow distribution with all samples densely clustered around 0. In contrast, sample-wise normalization results in a relatively wide-spread distribution, although it doesn't provide any indication of the actual flow size. 

For FlowGen, sample-wise normalization is more effective as the loss is directly calculated on the flows. We empirically found that avoiding excessively small flow scales leads to more stable loss, faster convergence, and more robust outputs. For FlowDiffusion, fixed size normalization proves to be more stable as the model benefits from the knowledge of actual scale. Using per-sample normalization for FlowDiffusion sometimes result in movements that are too large or too small. During inference, we normalize the sparse flow per sample, rescale the output dense flow using the maximum absolute magnitude of the input sparse drag instruction, and apply fixed size normalization before feeding it into the diffusion model. These normalization processes enable the diffusion model to generate good results without a specialized encoder network for optical flow.
\begin{figure*}
    \includegraphics[width=\linewidth]{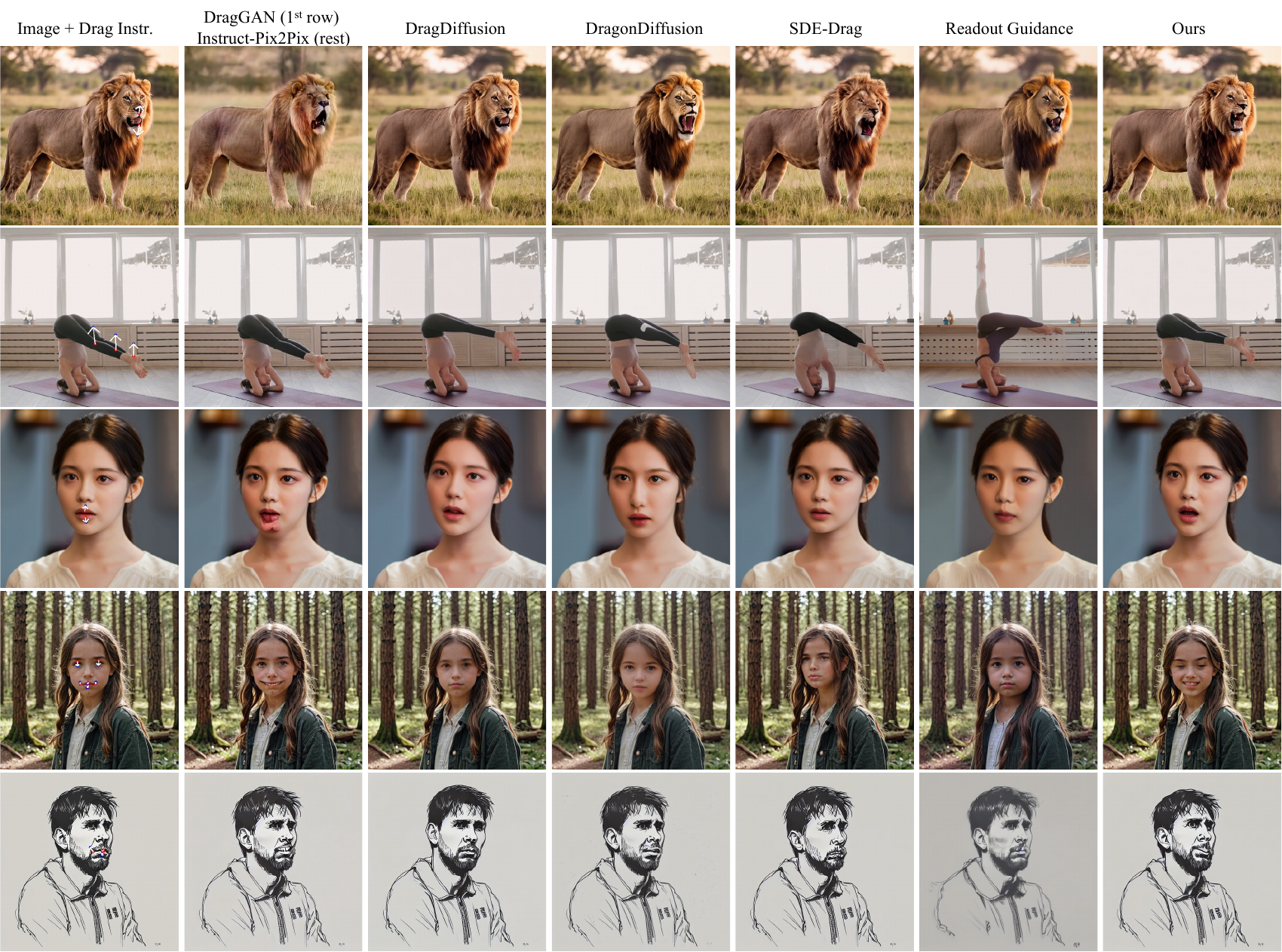}
    \vspace{-5mm}
    \caption{Qualitative comparison with other models. Zoom in for finer image details and drag instructions. (Second row input: \textcircled{c} V. Karpovich, via Pexels)}
    \label{fig:Qualitative}
    \vspace{-3mm}
\end{figure*}
\section{Experiments}
\label{sec:4}
\subsection{Settings}
\label{sec:4.1}
For face editing, we train our models on CelebV-Text~\cite{yu2023celebv}, a large-scale dataset of 70,000 high-quality video clips from the web. We extract frames at 10fps and use the sliding window technique to sample frame pairs. We use FlowFormer for optical flow estimation and YOLO~\cite{redmon2016you} for fast mask generation. After filtering out noisy samples, we obtain a dataset of 8M pairs. We analyze model trends and characteristics in this dataset, including ablation experiments on the dataset size. For general scene editing, due to the aforementioned computational limitation, we opt for a two-stage strategy where the model is fine-tuned on short videos (10\(\sim\)60s) containing scenes and motions relevant to the input image.

For the user study, we sample 22 dragged results from various domains, including cartoons, drawings, and real-world images, with 12 being facial images and 10 being general scenes. Although the recently introduced DragBench~\cite{shi2023dragdiffusion} includes images from 10 general categories, we opt for alternative data since our approach requires short video clips for general scenes. Therefore, we perform additional qualitative evaluations on face manipulation following DragGAN and DragonDiffusion. To obtain ground truth for comparison, we use frames from the validation set of TalkingHead-1KH~\cite{wang2021one} and report PSNR, SSIM, LPIPS~\cite{zhang2018unreasonable}, and CLIP image similarity scores~\cite{radford2021learning} using 68 dlib keypoint-based editing.

The final number of parameters for FlowGen generator is 54M and FlowDiffusion is 860M. We perform all experiments on A6000 GPUs. We train FlowGen at 256px for approximately three days using a single GPU and train FlowDiffusion at 512px for roughly five days using 8 GPUs. Fine-tuning FlowGen and FlowDiffusion on a single video takes about 20 minutes. For sampling, we use DPM++ sampler~\cite{lu2022dpmpp} with 20 steps. For comparisons, we use images generated at 512px from authors' official implementations with default settings. More details on dataset generation and training are elaborated in the Appendix.
\begin{table*}[h!]
    \centering
    \caption{Comparison of different drag editing methods using TalkingHead-1KH data. We mark the best and second-best scores in bold and underlined numbers. O and E refer to scores calculated using the original and edited images, respectively. * denotes results taken from the original paper.}
    \label{tab:benchmark}
    \vspace{-2.5mm}
    \begin{tabular*}{\textwidth}{l@{\extracolsep{\fill}}cccccccccccc}
        \toprule
         & \multirow{2}{*}{Input} & \multirow{2}{*}{Time (s)} & \multirow{2}{*}{Mem. (GB)} & \multicolumn{2}{c}{PSNR ($\uparrow$)} & \multicolumn{2}{c}{SSIM ($\uparrow$)} & \multicolumn{2}{c}{LPIPS ($\downarrow$)} & \multicolumn{2}{c}{CLIP$^{img}$ ($\uparrow$)} \\ 
          \cmidrule(r){5-6} \cmidrule(r){7-8} \cmidrule(r){9-10} \cmidrule(r){11-12}
         &  &   &  & O & E & O & E & O & E & O & E \\ 
        \midrule
        DragDiffusion & \multicolumn{1}{l}{Image, Drag, Prompt, Mask} & 75.3 & 11.6 & 23.83 & \textbf{23.59} & 0.81 & \textbf{0.77} & \underline{0.194} & \textbf{0.216} & \textbf{0.957} & \underline{0.945} \\ 
        DragonDiffusion & \multicolumn{1}{l}{Image, Drag, Prompt, Mask} & \underline{11.0} & \underline{6.6} & 21.53 & 21.26 & 0.80 & 0.74 & 0.233 & 0.260 & 0.895 & 0.891 \\ 
        SDE-Drag & \multicolumn{1}{l}{Image, Drag, Prompt, Mask} & 53.0 & 7.4 & 15.71 & 15.40 & 0.68 & 0.61 & 0.347 & 0.389 & 0.666 & 0.665 \\ 
        Readout Guidance & \multicolumn{1}{l}{\underline{Image, Drag, Prompt}} & $55.4^{*}$ & $19.2^{*}$ & \underline{25.44} & 21.26 & \textbf{0.87} & 0.71 & 0.205 & 0.289 & 0.892 & 0.885 \\ 
        InstantDrag (Ours) & \multicolumn{1}{l}{\bf{Image, Drag}} & \textbf{1.1} & \textbf{3.4} & \textbf{26.51} & \underline{22.92} & \underline{0.85} & \underline{0.75} & \textbf{0.154} & \underline{0.224} & \textbf{0.957} & \textbf{0.948} \\
        \bottomrule
    \end{tabular*}
    \vspace{-2mm}
\end{table*}

\subsection{Qualitative Evaluation}
\label{sec:4.2}
We compare our method with other recent approaches. For DragGAN, as stated in~\cite{zhang2024gooddrag}, we observe that PTI inversion~\cite{roich2022pivotal} of real images and subsequent editing shows suboptimal performance, even with minor distribution shifts. Due to these difficulties, we include DragGAN only with the lion sample in Fig.~\ref{fig:Qualitative} and provide the Instruct-Pix2Pix baseline, using text descriptions to move the object. 

\begin{figure}
    \includegraphics[width=0.995\linewidth]{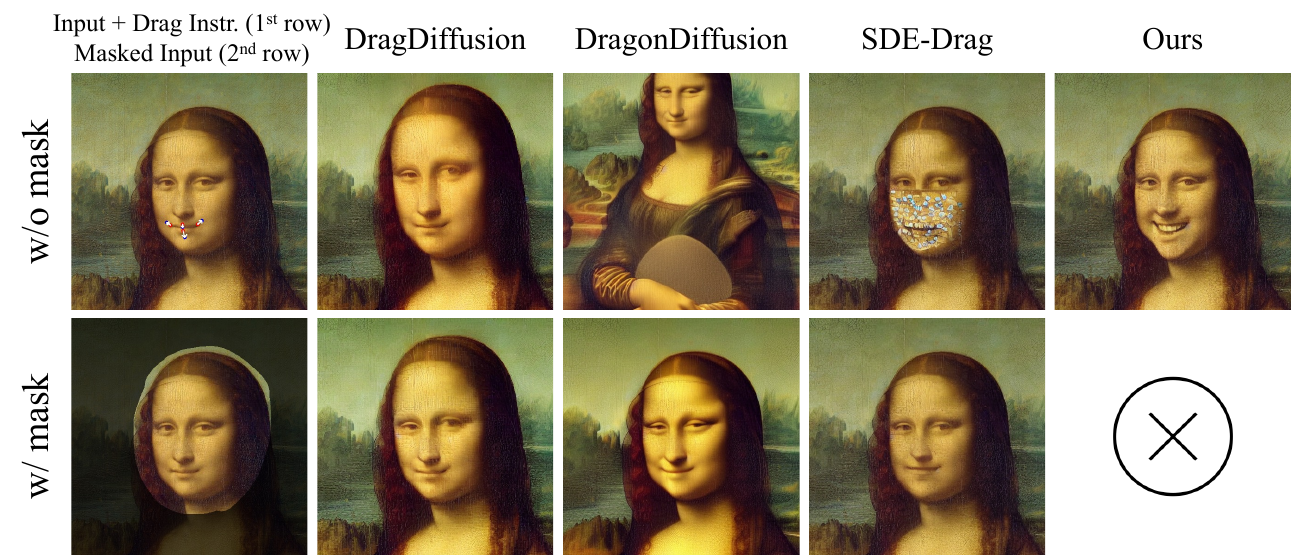}
    \vspace{-5mm}
    \caption{Generated samples with and without masking. As an inversion-free method, InstantDrag produces consistent drag edits without using a mask. (Input: \textcircled{c} Louvre Museum\textsuperscript{\ref{fn:louvre2}})}
    \label{fig:figureonly_usingmask}
\end{figure}

\begin{figure}
    \includegraphics[width=0.995\linewidth]{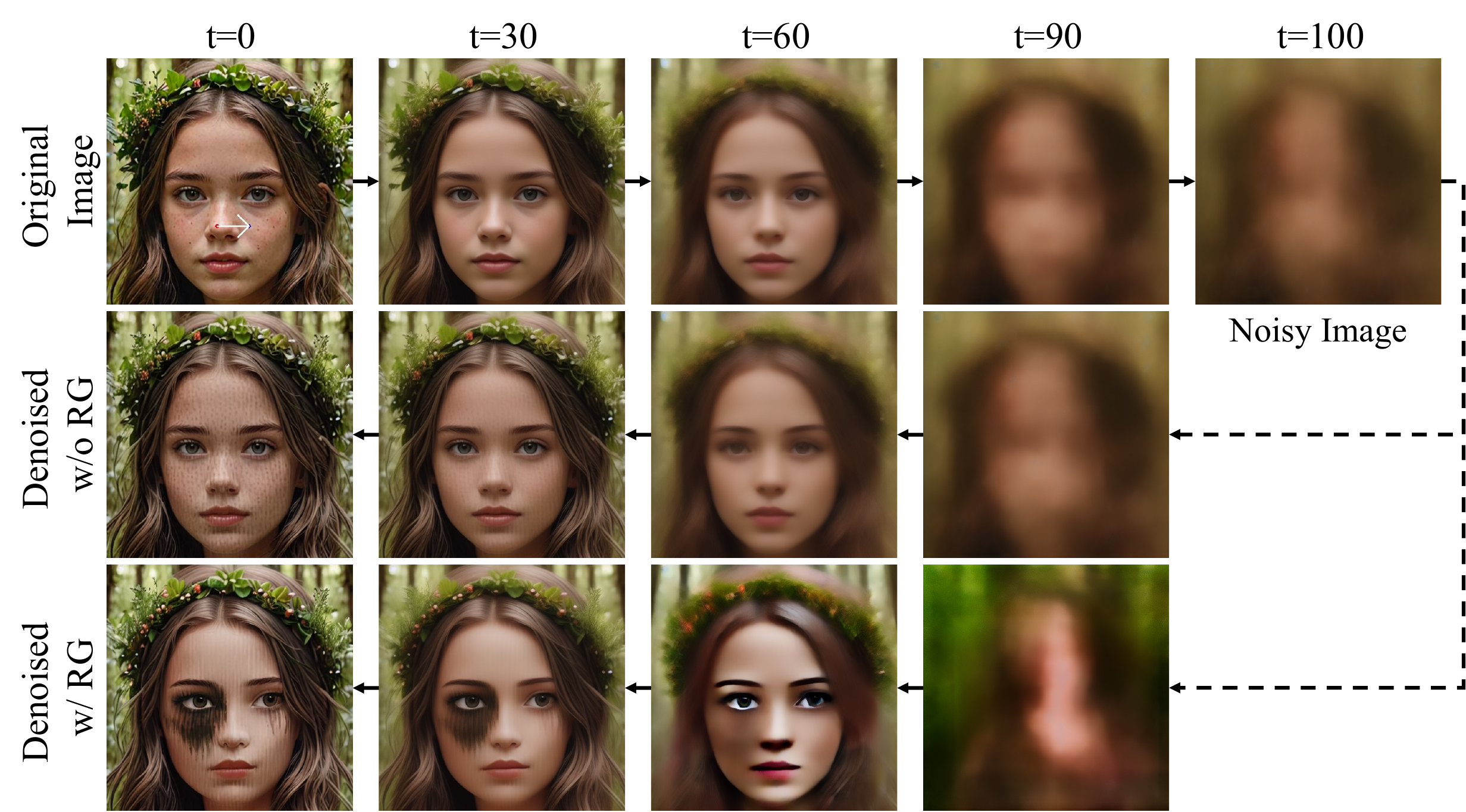}
    \vspace{-5mm}
    \caption{DDIM inversion and subsequent denoising process (100 step) in Readout Guidance (RG). We observe that fine details are often lost during inversion. Additionally, guidance-based methods can be sensitive to prompts and may exhibit some inconsistency.}
    \label{fig:figureonly_inversion}
\end{figure}

The first observation is about the preservation of high-frequency features. Most methods employ DDIM inversion to convert the real image. However, as shown in Fig.~\ref{fig:Qualitative} and Fig.~\ref{fig:figureonly_usingmask}, this process often results in a significant loss of detailed information. Without explicitly overfitting a network using techniques like LoRA or utilizing masks to set fixed regions, na\"ively guiding with text can lead to a change in identity. As can be seen in Fig.~\ref{fig:figureonly_inversion}, this issue is particularly noticeable for methods like Readout Guidance, where neither LoRA training nor masking is applied. Our method, being inversion-free, excels at preserving fine details even without the use of masks. 

\begin{figure*}
    \includegraphics[width=0.96\linewidth]{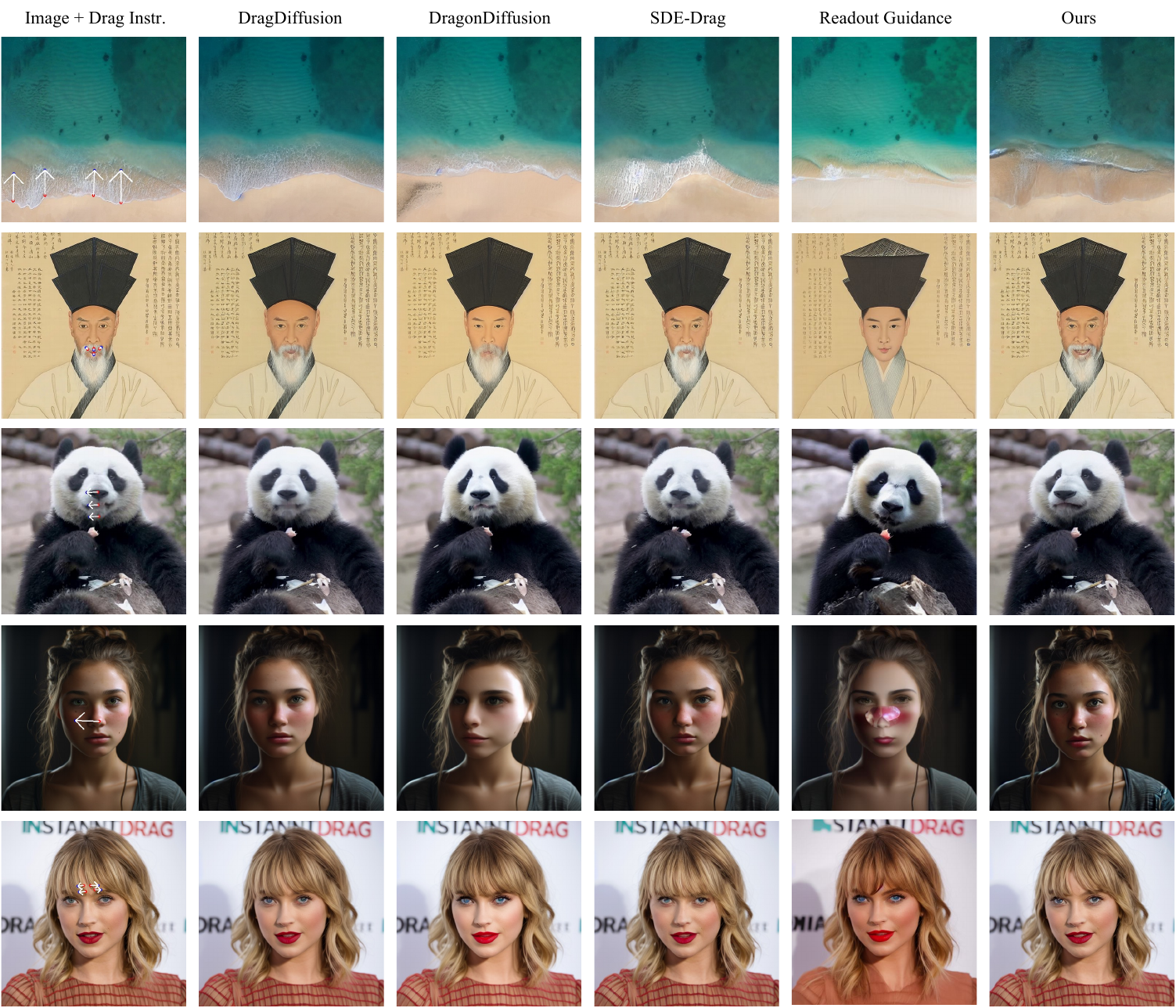}
    \vspace{-2mm}
    \caption{Additional qualitative comparison with other models. 1$\sim$4th row inputs: \textcircled{c} J. Loiterton (Pexels), \textcircled{c} National Museum of Korea\textsuperscript{\ref{fn:nmk}}, \textcircled{c} mds524680 (Pixabay), and DragBench \textcircled{c}~\cite{shi2023dragdiffusion}.}
    \label{fig:figureonly_quali}
\end{figure*}

The second observation is the exceptional generalization ability of our model. Although trained solely on real-world facial videos, our method effectively generalizes to domains like drawings and cartoons as shown in Fig.~\ref{fig:teaser},~\ref{fig:Qualitative}, and~\ref{fig:figureonly_quali}. Interestingly, though in preliminary stages, our model also demonstrates capability in handling non-facial general objects and scenes without fine-tuning. For a more in-depth discussion on the generalizability of our model, please refer to the Appendix.

Lastly, we observe that our model performs well in generating fine-grained movements, such as facial expressions, as shown in Fig.~\ref{fig:teaser},~\ref{fig:Qualitative} and~\ref{fig:figureonly_quali}. We attribute this to FlowGen's capability to accurately generate pixel-level motion cues, which efficiently guides the denoising process of FlowDiffusion.

\begin{figure}[t]
    \includegraphics[width=0.875\linewidth]{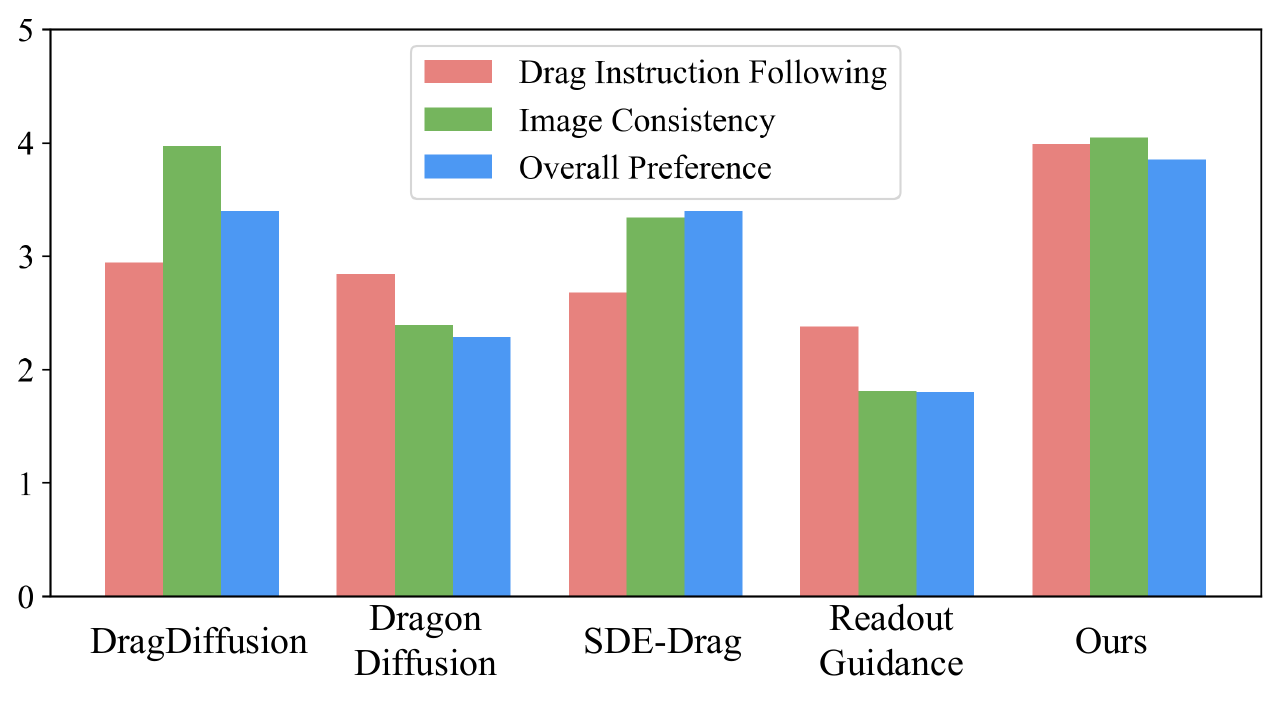}
    \vspace{-2mm}
    \caption{Bar plots of human evaluation results.}
    \label{fig:user_study}
    \vspace{-3.5mm}
\end{figure}

\begin{figure*}
    \includegraphics[width=\linewidth]{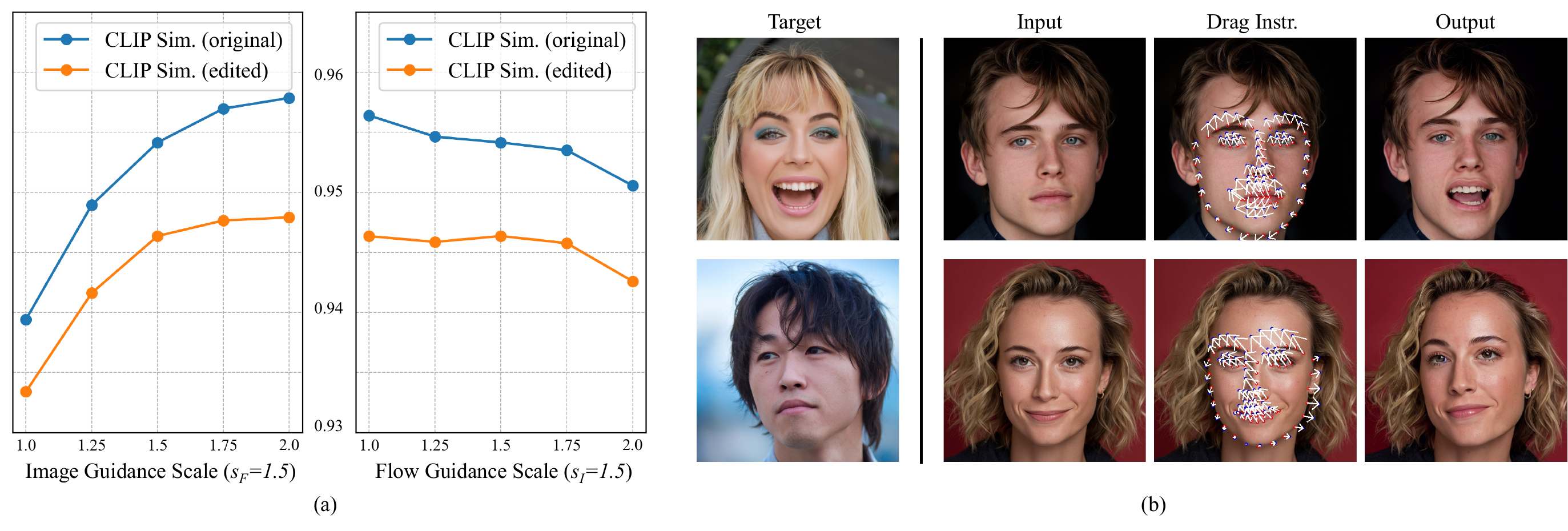}
    \vspace{-5mm}
    \caption{(a) Ablation studies on image and flow guidance using a subset of TalkingHead-1KH. Increasing image guidance makes generated images more like the original, while increasing flow guidance causes them to diverse from the original. Another scale is fixed at 1.5. (b) Results for face landmark manipulation.}
\label{fig:figureonly_ablationandtransfer}
\end{figure*}

\subsection{Quantitative Evaluation}
\label{sec:4.3}
Fig.~\ref{fig:user_study} presents human evaluation results based on 66 responses. Participants rated each model's output on a scale from 1 (very poor) to 5 (very good) across three criteria: instruction-following, identity preservation, and overall preference. The overall preference rating considered both instruction-following and identity preservation comprehensively, reflecting participants' inclination to use each model for drag-based image editing tasks.

Even without considering the advantage of our models being additionally fine-tuned on short videos for general scenes, they excel in zero-shot facial editing and overall preference. We attribute this to our model's strong ability to preserve high-frequency features and generate fine-grained movements. By learning motion cues from real-world data, our model generates plausible images that balance instruction-following and consistency, which is particularly important for human faces, where people are highly sensitive to subtle differences.

Table~\ref{tab:benchmark} presents the evaluation results on 100 frame pairs from the TalkingHead-1KH dataset. We generate drag instructions using 68 dlib keypoints between two sampled frame pairs, serving as the original and oracle edited images. For models requiring masks or text prompts, we manually annotate masks and generate captions using LLaVA~\cite{liu2024visual}. While Mean Distance (MD) has been recently used, we could not employ it due to noisy or out-of-distribution outputs from some models, which hindered keypoint detection. Moreover, we observed cases where models generated implausible images that lacked consistency and realism, merely moving keypoints without considering the context. To address these issues, we report results calculated with respect to both the original and edited images, with the latter representing the ground truth movement. Scores calculated against original images indicate content preservation, while those calculated against edited images reflect the resemblance to actual ground truth movement. Quantitative scores and qualitative inspection suggest that DragDiffusion and InstantDrag generally outperform other models. Notably, DragDiffusion, a point tracking and optimization-based method, accurately moves points to exact locations, whereas our model prioritizes consistency and plausible movements. 

We additionally perform ablation experiments to validate the effect of image and flow guidance scales as shown in Fig.~\ref{fig:figureonly_ablationandtransfer}.

\subsection{Discussion and Limitation}
\label{sec:4.4}
After a thorough evaluation, we identify the following strengths of our method: 1) it excels at preserving consistency, especially high-frequency features, even without a mask, due to being an inversion-free method; 2) it generates plausible images with realistic motions; and 3) it offers lighter and more efficient model pipeline and increased interactivity due to requiring fewer user inputs. Nevertheless, we also note a few limitations. Since our model's learning is based on the output of an optical flow estimation network, it faces difficulty handling very large motions that optical flow networks cannot capture. Additionally, while our model shows promising generalizability (as detailed in the Appendix), we occasionally observe limitations in preserving identity or creating accurate motions for non-facial scenes without fine-tuning, due to the model being trained solely on facial videos. However, given the positive signs of generalizability, we believe training on diverse motions from larger datasets can further improve performance across various domains.

\begin{figure*}
    \includegraphics[width=0.96\linewidth]{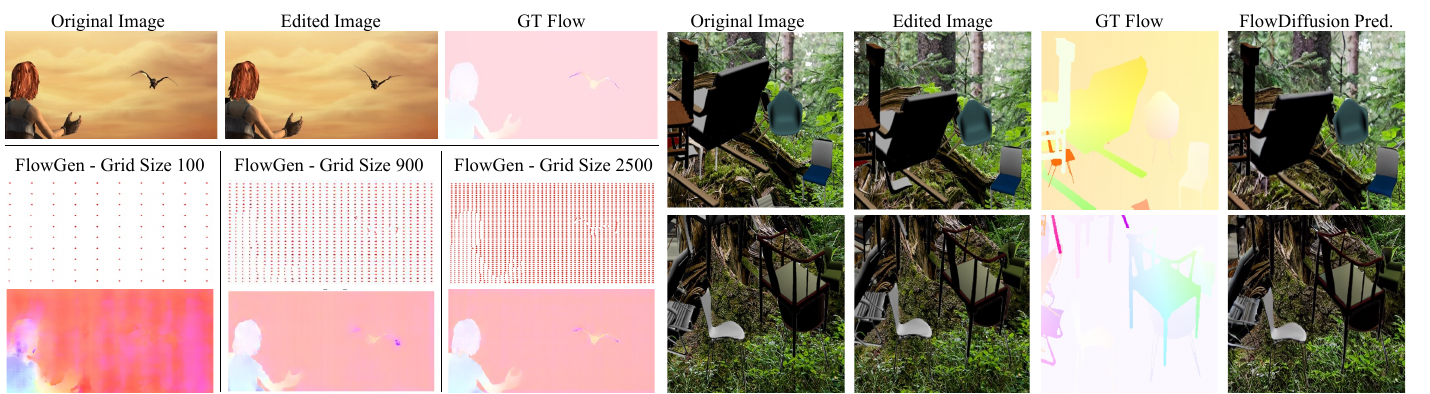}
    \vspace{-2mm}
    \caption{FlowGen predictions for SINTEL dataset (left) with varying grid size. FlowDiffusion predictions for FlyingChairs dataset (right). Samples from SINTEL dataset \textcircled{c}~\cite{butler2012naturalistic}, and FlyingChairs dataset \textcircled{c}~\cite{dosovitskiy2015flownet}.}
    \label{fig:figureonly_flowgen_flowdiffusion}
\end{figure*}

\section{Conclusion}
\label{sec:5}
While large-scale foundational models are being distilled to generate high-quality images from texts within a second, adding precise controls to generated images still lags behind. Inspired by the recent drag editing paradigm, we proposed InstantDrag, a drag-dedicated pipeline composed of carefully designed and trained FlowGen and FlowDiffusion. Our method not only achieves faster speeds and a lighter model but also enhances interactivity by reducing inputs to just an image and a drag instruction. This potentially decreases the editing time even further. Given the intuitive nature and interactivity of drag editing, we believe our work points in a promising direction for improving the accessibility and efficiency of real-time interactive editing on a wide range of devices.

\begin{acks}
This work was supported by the National Research Foundation (NRF) grant (RS-2024-00405857, 95\%) and the IITP grant (RS-2021-II211343: AI Graduate School Program, Seoul National University, 5\%) funded by the Korean government (MSIT).
\end{acks}

\footnotetext[2]{\label{fn:louvre2}\textcircled{c} 2011 GrandPalaisRmn  (musée du Louvre) / Michel Urtado. Mona Lisa. Louvre Collections: https://collections.louvre.fr/ark:/53355/cl010062370}
\footnotetext[3]{\label{fn:nmk}\textcircled{c} National Museum of Korea. Treasure Lee Chae Portrait (2006). Public Nuri, KOGL Type 1. Link: https://www.museum.go.kr/site/main/relic/search/view?relicId=1138}


\bibliographystyle{ACM-Reference-Format}
\bibliography{siggraph_a}
\nobalance

\clearpage
\appendix
\setcounter{table}{0}
\setcounter{page}{1}
\setcounter{figure}{0}
\setcounter{equation}{0}
\renewcommand{\thefigure}{A\arabic{figure}}
\renewcommand{\thetable}{A\arabic{table}}
\section{On the Generalizability of the InstantDrag Pipeline}
\label{sec:sup_generalizability}
As discussed in the main paper, our base models (FlowGen and FlowDiffusion) are trained exclusively on the facial video dataset CelebV-Text~\cite{yu2023celebv}. To handle general scene editing, we proposed a fine-tuning approach for FlowGen and FlowDiffusion using short videos containing relevant scenes. Interestingly, we observed that our base (non-fine-tuned) model successfully generalizes to unseen (non-facial) general scenes and objects in many cases. We hypothesize that this generalization capability stems from training on relatively in-the-wild, uncurated videos, allowing our models to naturally learn some basic dynamics of real-world objects.

\begin{figure*}
    \includegraphics[width=0.89\linewidth]{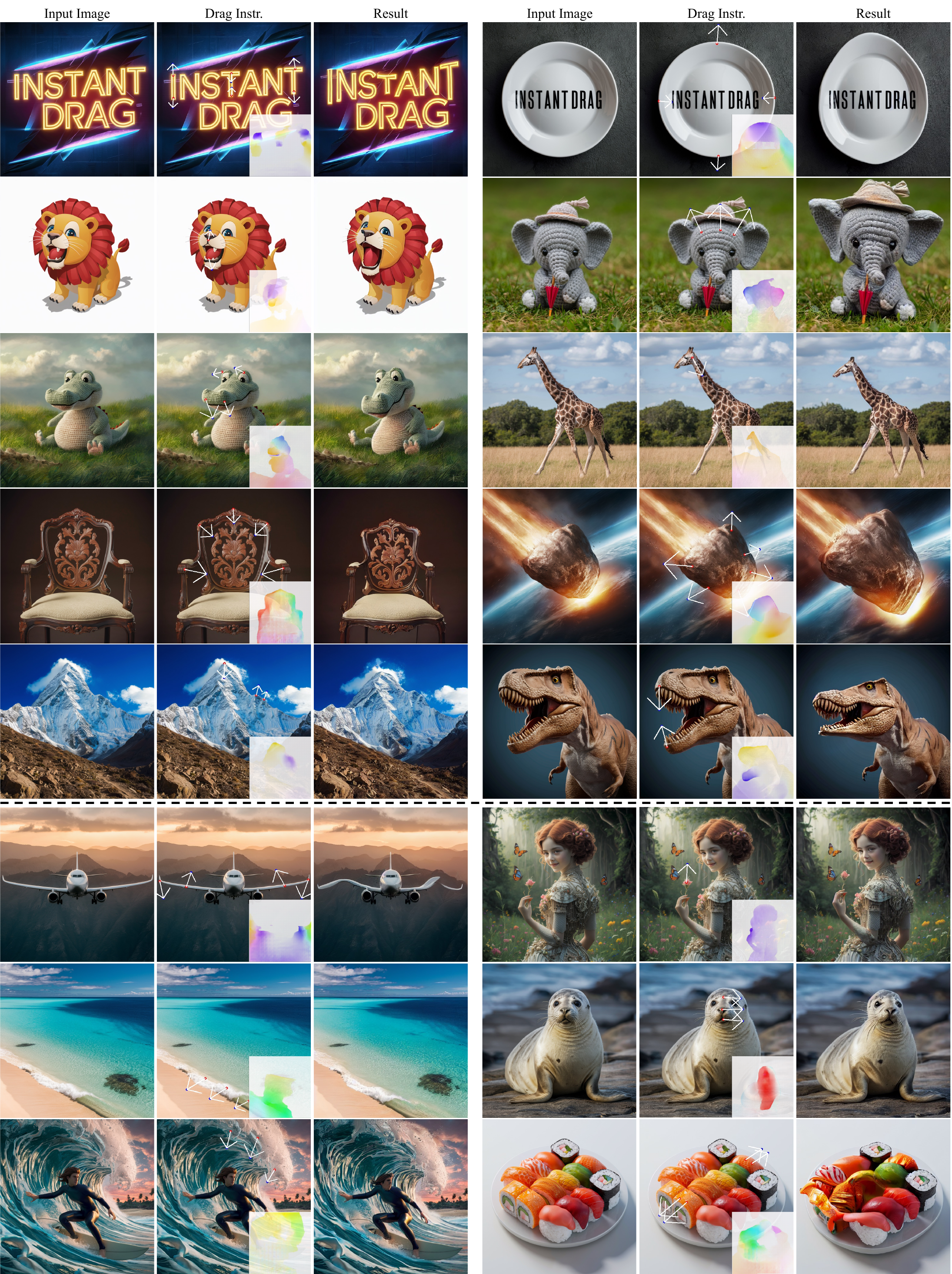}
    \vspace{-2mm}
    \caption{Qualitative results on editing general scenes and objects using our base model without fine-tuning.}
    \label{fig:sup_generalizability}
\end{figure*}

Fig.~\ref{fig:sup_generalizability} presents editing results of non-facial images. The first five rows showcase relatively successful results, while the last three rows demonstrate less successful outcomes. We observe that many movements successfully generalize to completely unseen domains such as calligraphy, 3D objects, animals, and natural scenes. However, since our models were not explicitly trained on such diverse data, they sometimes struggle to preserve subtle details or produce accurate motions in complex images.

Typical failure cases include: 1) instances where images are not properly dragged or moved, resulting in outputs nearly identical to the input, 2) undesired movements, such as shifting the entire background or the whole object instead of just a specific part, and 3) cases where fine details are lost or undesired artifacts are produced. Our analysis reveals that issues 1) and 2) typically occur when FlowGen's output is inaccurate. Given that FlowGen is trained solely on facial videos, it sometimes fails to distinguish between parts that should be moved and those that should remain static. This ambiguity often leads to inaccurate motions or minimal movement. Issue 3), the loss of details or production of artifacts, is primarily attributed to the capacity limitations of FlowDiffusion. As it is also trained exclusively on facial videos, it sometimes struggles to accurately preserve fine details in completely unseen domains. For extremely fine-grained images with intricate details, we suspect that this issue is also related to the capacity limitations of Stable Diffusion v1.5~\cite{rombach2022high}, which serves as the backbone for our FlowDiffusion model.

While our models show reasonable results without fine-tuning, the general scene editing results presented in the main paper are based on a fine-tuning approach using base models trained on SINTEL~\cite{dosovitskiy2015flownet} and FlyingChairs datasets~\cite{butler2012naturalistic}. We observe similar results when initializing the fine-tuning process with our base CelebV-trained FlowGen and FlowDiffusion models instead of those trained on SINTEL and FlyingChairs. Although both models can be fine-tuned end-to-end, we find that applying LoRA fine-tuning (rank 8) to FlowDiffusion yields comparable results with minimal storage overhead. Following the approach in~\cite{luo2023lcm}, we add low-rank matrices to selected convolutional and attention layers. For fine-tuning scenarios, we use short videos (10-60 seconds) capturing the motion of the target object. We process these videos using the same methodology as for CelebV-Text, except that we employ Grounded-SAM to obtain binary masks and use grid-based sampling for the extraction of pseudo-drag instructions. The fine-tuning process for FlowGen and FlowDiffusion on a single video takes approximately 20 minutes.

\section{Additional Evaluation Samples}
\begin{figure*}
    \includegraphics[width=0.93\linewidth]{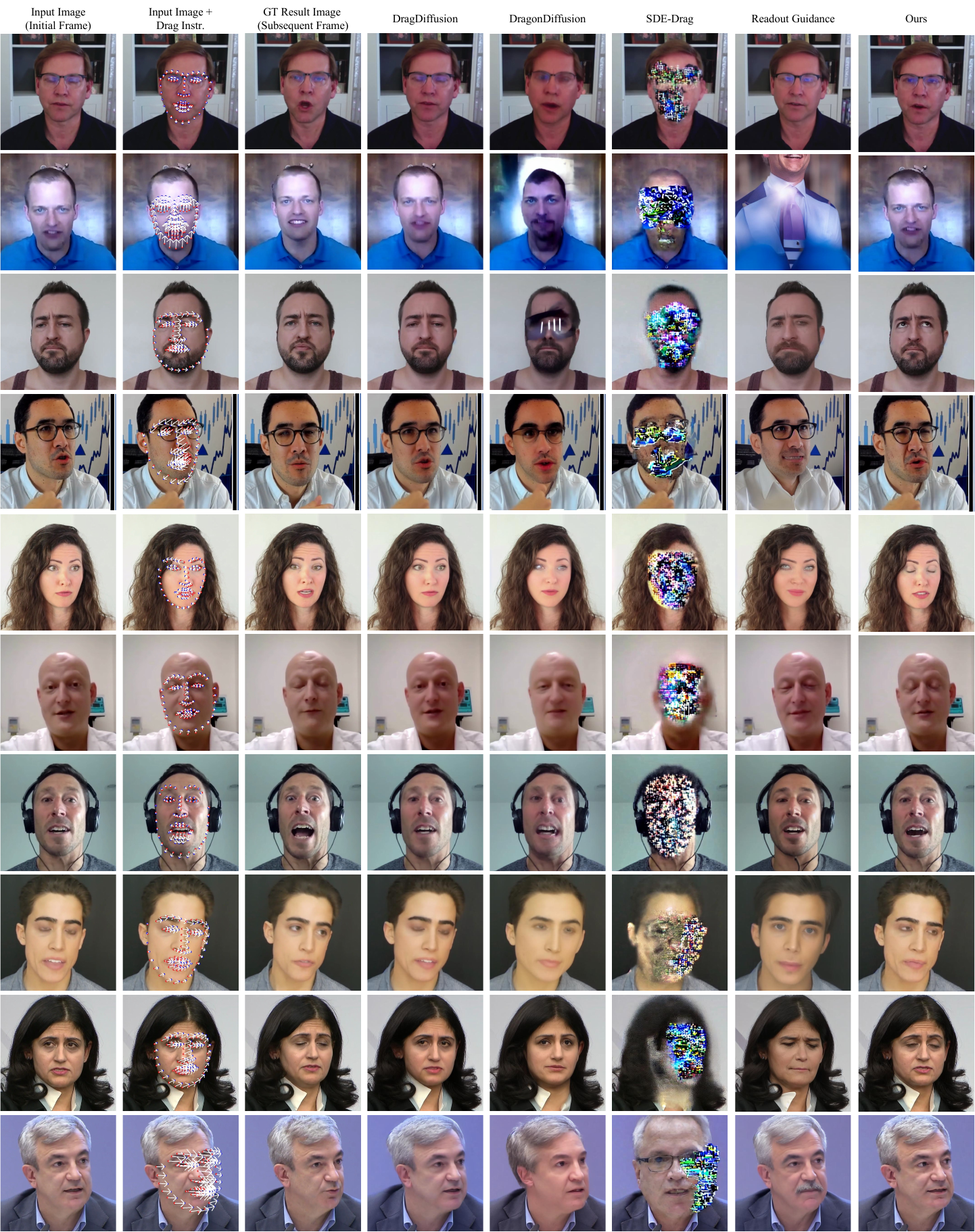}
    \caption{Additional qualitative results. (Inputs: TalkingHead-1KH \textcircled{c}~\cite{wang2021one})}
    \label{fig:sup_talkinghead}
\end{figure*}

\begin{figure*}
    \includegraphics[width=0.98\linewidth]{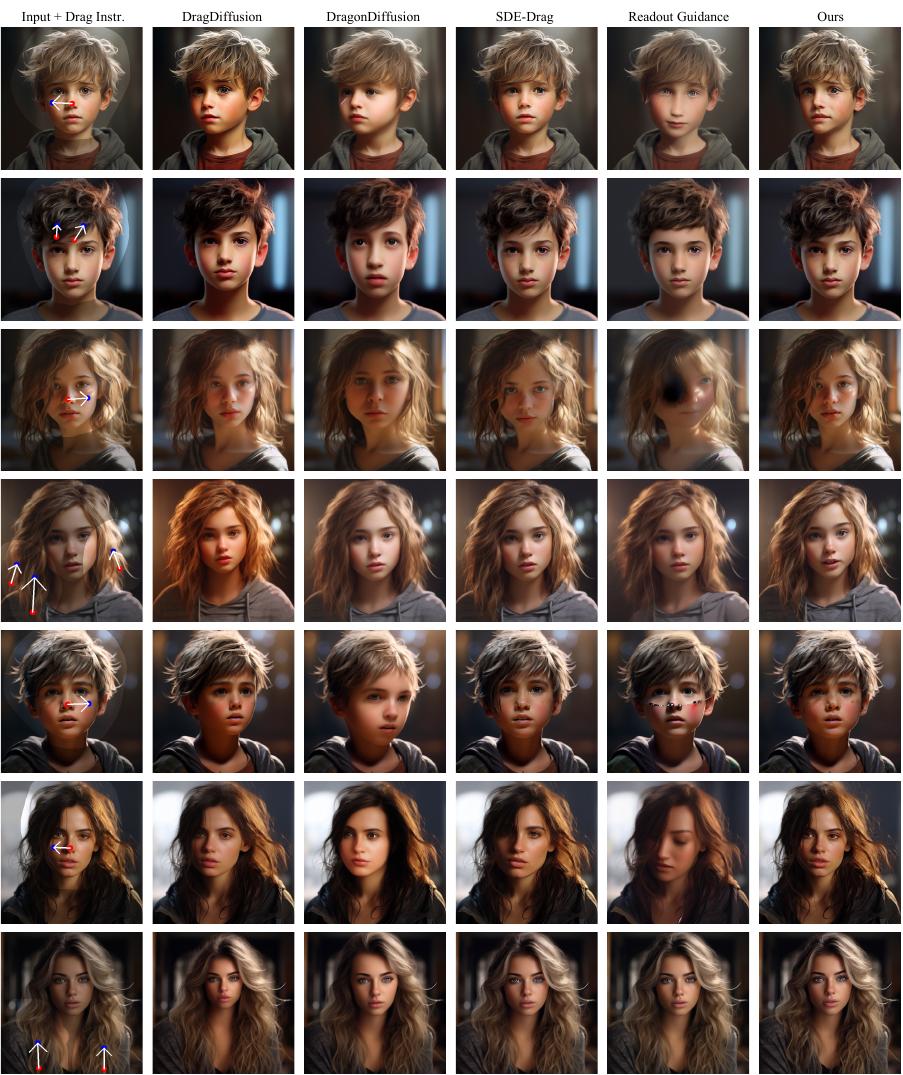}
    \caption{Additional qualitative results. (Inputs: facial images of DragBench \textcircled{c}~\cite{shi2023dragdiffusion})}
    \label{fig:sup_dragbench}
\end{figure*}

We additionally provide more qualitative results on our base model. Randomly selected qualitative results for TalkingHead-1KH~\cite{wang2021one} data used in Table~\ref{tab:benchmark} are provided in Fig.\ref{fig:sup_talkinghead}. As mentioned in the original paper, we guide the drag editing process using 68 dlib keypoints extracted from two video frames. Since TalkingHead-1KH includes some wild samples of low resolution and unaligned faces, we observe that the editing results sometimes yield significant artifacts or complete changes of identity, occasionally resulting in images that no longer resemble faces. As we are consistently manipulating 68 points, we used the third variant of our model presented in Sec.~\ref{sec:sup_pseudo_drag} with an image guidance scale of 1.75 and a flow guidance scale of 1.5. We also provide qualitative results on the facial images of DragBench~\cite{shi2023dragdiffusion} in Fig.~\ref{fig:sup_dragbench}, which mainly consists of editing scenarios involving fewer drag instructions on synthetic faces.

\section{Comparisons with Other Models at Different speeds}
\begin{figure*}
    \includegraphics[width=0.85\linewidth]{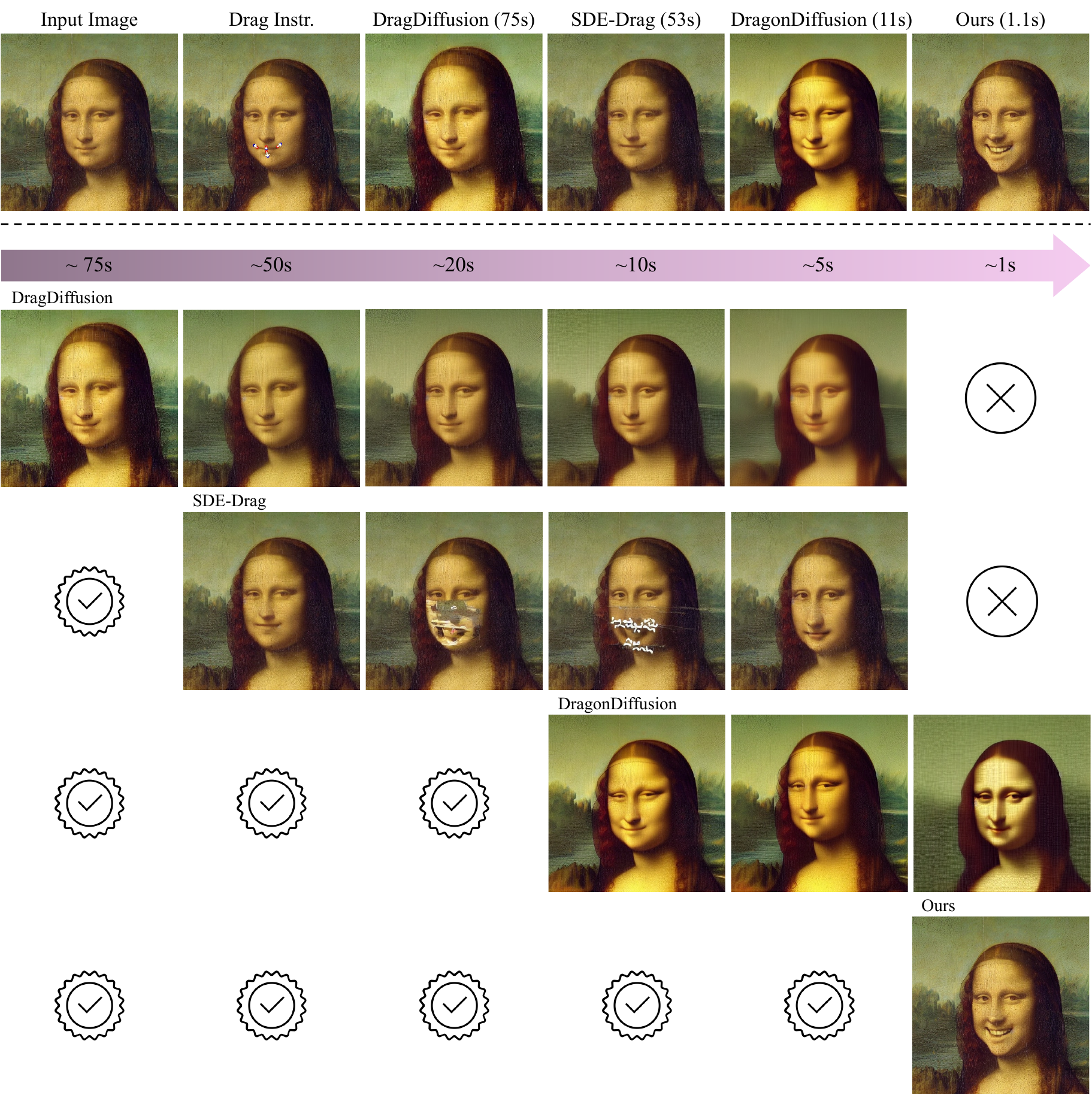}
    \caption{Comparisons with other models operating at different speeds. The specific conditions under which these images were generated are detailed in Table~\ref{tab:sup_othermodels}. A check and cross indicate that further extension or reduction in time is not necessary or feasible, respectively. (Inputs: \textcircled{c} Louvre Museum\textsuperscript{\ref{fn:louvre3}})}
    \label{fig:sup_othermodels}
    \vspace{-2mm}
\end{figure*}

We compare our method with other models operating at various speeds. To reduce inference times for models like DragDiffusion, we can adjust the number of LoRA training steps, latent optimization steps, and DDIM inversion and sampling steps. As there can be multiple possible combinations for step reduction to match specific time constraints, we use configurations which we found to be effective based on our heuristic trials. Qualitative results are presented in Fig.~\ref{fig:sup_othermodels}, with corresponding settings detailed in Table~\ref{tab:sup_othermodels}. In Fig.~\ref{fig:sup_othermodels}, a check and cross indicate that further extension or reduction in time for that particular model is not necessary or feasible, respectively.

All time-dependent experiments in our paper were conducted using a single A6000 GPU. We note that employing an industry-standard A100 GPU further reduces our model's latency by 35\% and achieves an impressive editing time of 0.72 seconds.

\begin{table}[h!]
    \centering
    \caption{Settings used for Fig.~\ref{fig:sup_othermodels}. A dash (--) indicates that the specific operation is not used for that method. Times measured with an A6000 gpu.}
    \label{tab:sup_othermodels}
    \begin{tabular}{@{}lccccc@{}}
        \toprule
         & Actual & LoRA & Latent & (Inversion) \& \\
         & Time (s) & steps & Opt. steps & Sampling steps \\ 
        \midrule
        DragDiffusion & \multirow{5}{*}{} &&&\\
        \quad $\sim$75s & 75.3 & 80 & 80 & 50 \\
        \quad $\sim$50s & 52.0 & 40 & 80 & 20 \\
        \quad $\sim$20s & 23.0 & 20 & 30 & 10 \\
        \quad $\sim$10s & 10.5 & 0 & 10 & 4 \\
        \quad $\sim$5s & 6.3 & 0 & 5 & 2 \\
        \addlinespace
        SDE-Drag & \multirow{4}{*}{} &&&\\
        \quad $\sim$50s & 53.0 & 100 & -- & 100 \\
        \quad $\sim$20s & 19.5 & 40 & -- & 40 \\
        \quad $\sim$10s & 9.0 & 0 & -- & 40 \\
        \quad $\sim$5s & 5.5 & 0 & -- & 20 \\
        \addlinespace
        DragonDiffusion & \multirow{3}{*}{} &&&\\
        \quad $\sim$10s & 11.0 & -- & -- & 50 \\
        \quad $\sim$5s & 4.67 & -- & -- & 20 \\
        \quad $\sim$1s & 0.98 & -- & -- & 3 \\
        \addlinespace
        Ours & \multirow{2}{*}{} &&&\\
        \quad $\sim$1s & 1.10 & -- & -- & 20 \\
        \bottomrule
    \end{tabular}
\end{table}

\section{Additional Details on Models}
This section provides additional descriptions regarding network design choices and extra details. Our trained models and codes will be open-sourced upon publication.

\subsection{FlowGen}
We viewed the task of generating dense optical flow from drag instructions as a translation task conditioned on an input image. To achieve this, we explored several one-step generative models. A notable consideration was the recently released Pix2Pix-Turbo \cite{parmar2024one}. Following Pix2Pix-Turbo, we experimented with fine-tuning a pre-trained SD-Turbo~\cite{sauer2023adversarial} by slightly modifying the input and output channels. We tested both full fine-tuning of SD-Turbo, including the autoencoder, and LoRA training of SD-Turbo as performed in Pix2Pix-Turbo. However, we observed that the training loss diverged under the settings we tried. This result indicates the difficulty of utilizing one-step text-to-image model priors for translation to a completely different motion domain. We believe that estimating scores for translation to a completely different domain (e.g., optical flow, normal map, etc.) in one-step generative models remains challenging and constitutes an important topic for future research. We found that training FlowGen with a U-shaped Generator and discriminator from scratch yielded acceptable results for our task.

FlowGen is expected to correctly interpret all input sparse flows and map them to dense optical flow. Since a single motion can be represented via multiple different sets of points, we sampled four different sparse flows using pseudo drag instruction as detailed in Sec.~\ref{sec:3.2.2}. We then updated the generator four times to map these different outputs to the same ground truth flow while updating the discriminator once. More details on extracting pseudo drag instructions can be found in \ref{sec:sup_pseudo_drag}.

Since our FlowGen generator is based on a U-shaped convolutional network, it can accept images of various resolutions. For training, we primarily used 256x256 images at a batch size of 256 and the Adam optimizer~\cite{KingBa15} with a constant learning rate of 2e-4. Given that we predict normalized flow values, our model scales well to images of larger or smaller sizes. For the SINTEL dataset, we resized images and trained our model on 512x256 images. Since FlowGen is lightweight, with 54 million parameters for the generator and 15 million for the discriminator, we performed full precision (fp32) training for 80,000 steps using a single A6000 GPU, which took approximately three days. For training FlowGen's generator, we employ an objective:
\begin{align}
    G^* = \arg \min_G \max_D \mathcal{L}_{adv}(G, D) + \lambda \mathcal{L}_{rec}(G),
\end{align}
where $\lambda=100$. $\mathcal{L}_{adv}(G,D)$ and $\mathcal{L}_{rec}(G)$ are as defined in the main paper (Sec.~\ref{sec:3.1.1}). The inference time of FlowGen in an actual drag editing scenario is just 0.003 seconds, which is marginal compared to FlowDiffusion.

\subsection{FlowDiffusion}
Given a denoising network \(\epsilon_\theta(z_t, c_I, c_T)\), a noisy latent \(z_t\) and input conditions \(c_I, c_T\), Instruct-Pix2pix performs classifier-free guidance on both the image and text domains using the modified score estimate as follows:
\begin{align}
    \tilde{\epsilon}_\theta(z_t, c_I, c_T) = & \ \epsilon_\theta(z_t, \varnothing, \varnothing) \nonumber \\
    & + s_I \cdot [\epsilon_\theta(z_t, c_I, \varnothing) - \epsilon_\theta(z_t, \varnothing, \varnothing)] \nonumber \\
    & + s_T \cdot [\epsilon_\theta(z_t, c_I, c_T) - \epsilon_\theta(z_t, c_I, \varnothing)].
\end{align}

Inspired by the effectiveness of this approach, we adapted the pre-trained Stable Diffusion v1.5's U-Net to accept flow conditions by introducing additional input channels, resulting in a denoising network, \(\epsilon_\theta(z_t, c_I, c_F, c_T)\). We began with a straightforward approach, utilizing 1.5 for $s_I$ and $s_F$ and 3.0 for $s_T$ unless otherwise noted.
\begin{align}
    \tilde{\epsilon}_\theta(z_t, c_I, c_F, c_T) = & \ \epsilon_\theta(z_t, \varnothing, \varnothing, \varnothing) \nonumber \\
    & + s_I \cdot [\epsilon_\theta(z_t, c_I, \varnothing, \varnothing) - \epsilon_\theta(z_t, \varnothing, \varnothing, \varnothing)] \nonumber\\
    & + s_F \cdot [\epsilon_\theta(z_t, c_I, c_F, \varnothing) - \epsilon_\theta(z_t, c_I, \varnothing, \varnothing)] \nonumber \\
    & + s_T \cdot [\epsilon_\theta(z_t, c_I, c_F, c_T) - \epsilon_\theta(z_t, c_I, c_F, \varnothing)].
    \label{eq:cfg1}
\end{align}
Since changes in the training pairs were less correlated with the text condition, we observed that the trained score estimator's predictions remained consistent regardless of the text input. Motivated by this observation, we tested two additional variants by explicitly disabling the text condition, consistently setting the input to a null embedding.
\begin{align}
    \tilde{\epsilon}_\theta(z_t, c_I, c_F, c_T=\varnothing) = & \ \epsilon_\theta(z_t, c_I, c_F, \varnothing) \nonumber \\
    & + s_F \cdot [\epsilon_\theta(z_t, c_I, c_F, \varnothing) - \epsilon_\theta(z_t, c_I, \varnothing, \varnothing)].
    \label{eq:cfg2}
\end{align}
\begin{align}
    \tilde{\epsilon}_\theta(z_t, c_I, c_F, c_T=\varnothing) = & \ \epsilon_\theta(z_t, \varnothing, \varnothing, \varnothing) \nonumber \\
    & + s_I \cdot [\epsilon_\theta(z_t, c_I, \varnothing, \varnothing) - \epsilon_\theta(z_t, \varnothing, \varnothing, \varnothing)] \nonumber \\
    & + s_F \cdot [\epsilon_\theta(z_t, c_I, c_F, \varnothing) - \epsilon_\theta(z_t, c_I, \varnothing, \varnothing)].
    \label{eq:cfg3}
\end{align}

For Eq.\ref{eq:cfg2}, which does not perform classifier-free guidance on the input image, we did not drop any image during training. However, the inability to enforce explicit image guidance resulted in identity and color tone changes. Finally, we arrived at Eq.\ref{eq:cfg3}, where we perform classifier-free guidance only on the image and flow. As shown in Fig.~\ref{fig:figureonly_ablationandtransfer}, this results in a controllable setting suitable for drag-based editing, where users can adjust the image guidance scale and flow guidance scale to find a balance between the amount of motion and identity preservation.

We also attempted to integrate perceptual losses such as LPIPS into the FlowDiffusion training to preserve identity better. However, calculating LPIPS in pixel space requires forward and backward passes through the VAE's decoder, introducing extra overhead in training time and memory usage. Although models converged slightly faster in terms of steps, we found that models without LPIPS were much more efficient in computing and time. We believe that adopting recent efficient implementations of LPIPS in latent space could be beneficial in the future~\cite{kang2024distilling}.

We initiated FlowDiffusion training from a pre-trained Stable Diffusion 1.5 model. As mentioned in the main text, we dropped images 5\% of the time and flow 10\% of the time during training, ensuring that images were never dropped when the flow was present. For training, we used 512x512 images at a batch size of 512 and the AdamW optimizer~\cite{loshchilov2017decoupled} with a constant learning rate of 1e-4. The resulting FlowDiffusion model has roughly the same number of parameters as Stable Diffusion 1.5 (860M). We performed mixed precision training for 50,000 steps using 8 A6000 GPUs, which took approximately five days. We used fp16 for inference, and it takes about 1.1 seconds to sample 20 steps with the DPM++ sampler.

\begin{figure}[t]
    \centering  
    \includegraphics[width=\linewidth]{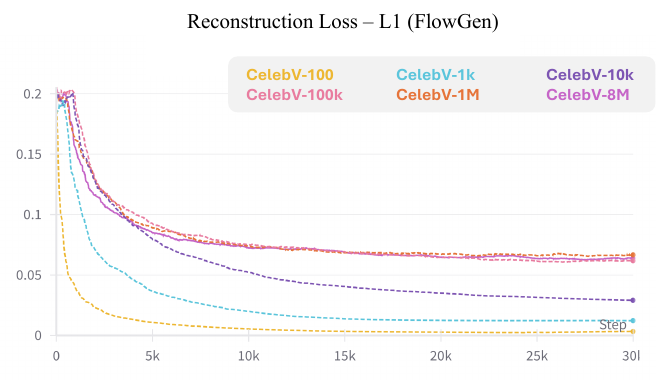}
    \caption{Reconstruction loss for FlowGen using different dataset sizes.}
    \vspace{-2mm}
    \label{fig:supdataset_ablation_flowgen}
\end{figure}

\subsection{Details on extracting pseudo drag instructions}
\label{sec:sup_pseudo_drag}
Here, we provide additional details on the random sparse flow sampling algorithm previously mentioned in the main text. When a mask is not available (\textit{e.g.}, in general scenes such as FlyingChairs), we use grid-based sampling instead of stochastic random sampling of target points in the mask regions.

We first initialize the sparse flow \(f_s \in \mathbb{R}^{2\times h\times w}\) with random values sampled from \(U(0,1)\). We also explore a variant where the sparse flow is initialized with random values proportional to the absolute magnitude of movement by multiplying the actual dense optical flow. The intuition here is to sample more from areas with significant movement, though this could lead to saturation of sampled points in specific locations. This approach is sometimes useful when a mask is not provided.

After initialization, we assign \(-\infty\) values to masked regions while adding a value \(v_g\) to grid locations and \(v_s\) to specific points, such as facial keypoints. Finally, we select the top-\(k\) points in the sparse flow and zero out the remaining parts, where \(k\) is randomly chosen within predefined ranges. For training, we introduce a \texttt{max\_points} parameter that determines the range of \(k\).

For the face model, we provide four variants:
\begin{enumerate}
    \item A model trained with a grid size of 100 and \texttt{max\_points} of 10, always including the nose coordinate as a specific point. (\(v_g = 0.4\), \(v_s = 1.0\))
    \item A model trained with a grid size of 100 and \texttt{max\_points} of 50, providing 23 key locations on the human face. (\(v_g = 0.4\), \(v_s = 0.7\))
    \item A model trained with a grid size of 100 without any specific points. (\(v_g = 0.4\), \(v_s = 0.0\))
    \item A model trained with grid size 900 without any specific points. (\(v_g = 0.4\), \(v_s = 0.0\))
\end{enumerate}
As shown in Sec.~\ref{sec:3.2.2} and Fig.~\ref{fig:sparse_flow}, we found the second variant (the stochastic model with 23 dlib keypoints) to be the most robust and performant overall. Unless otherwise specified, this variant served as our primary and base model for the figures presented. However, each variant excels in different scenarios. The first model (with nose coordinate) demonstrates particular strength in tasks involving single-point movements. In contrast, the third and fourth models (without specific points) excel in fine-grained editing tasks that require numerous control points or extremely precise and localized movements.

As previously emphasized, the sampling of appropriate points for FlowGen is crucial to its performance. While our stochastic sampling strategy based on grid and pre-defined specific points (facial keypoints) has proven effective, more sophisticated methods may further enhance the process. For instance, we can utilize the watershed strategy from~\cite{zhan2019self} to efficiently sample points along the edges of moving objects. This approach involves detecting motion edges with a Sobel filter, generating a topological-distance watershed map, and refining the key points through non-maximum suppression. Exploring these advanced sampling techniques remains a promising direction for future work.

\begin{figure}[t]
    \centering  
    \includegraphics[width=\linewidth]{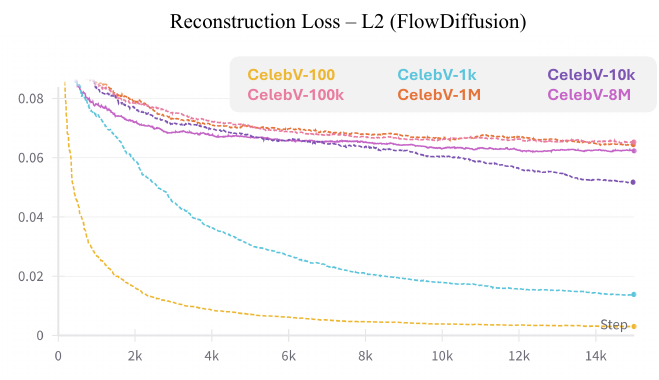}
    \caption{Reconstruction loss for FlowDiffusion using different dataset sizes.}
    \vspace{-2mm}
    \label{fig:supdataset_ablation_flowdiffusion}
\end{figure}

\section{Datasets}
Lastly, we provide details on curating the dataset. The most crucial part of dataset pre-processing is accurately masking and estimating the flow of the target object of interest. As long as we can accurately detect and mask the target object, new appearances of other objects are acceptable, as we enforce background consistency as mentioned in Sec.~\ref{sec:3.2.3}. However, maintaining consistent masks is challenging since we use YOLO to make human masks, which detect every part of a human. For instance, if a hand appears suddenly in the second frame, YOLO would detect and mask the hand in that frame but not in the first frame, leading to inconsistent mask data between the two frames. This issue is relatively minor in general scenes, as we are less sensitive to minor occlusions and subtle changes. However, the sudden appearance of large hands occluding or partially obscuring human faces can significantly alter the perception of these faces. To address this, we first filter out low-quality (low bit rate) videos and sample pairs. Among the sampled pairs, we use an additional YOLO-based hand detector and disregard samples where the number of hands or people changes. Using these curation methods, we obtain a total of 8 million pairs. To reduce storage overhead, we quantize optical flow into uint8 and de-quantize them on the fly during training, as saving optical flow in full fp32 precision consumes too much storage. In the case of general scenes where YOLO could not mask the object, we used GroundedSAM~\cite{ren2024grounded}.

\footnotetext[4]{\label{fn:louvre3}\textcircled{c} 2011 GrandPalaisRmn  (musée du Louvre) / Michel Urtado. Mona Lisa. Louvre Collections: https://collections.louvre.fr/ark:/53355/cl010062370}

We ablate the number of pairs required to train FlowGen and FlowDiffusion. Using CelebV-Text's 8 million pairs, we create six subsets: CelebV-Text-100, CelebV-Text-1k, CelebV-Text-10k, CelebV-Text-100k, CelebV-Text-1M, and the original CelebV-Text-8M. We train FlowGen and FlowDiffusion for a short duration (30k and 15k steps, respectively) and empirically validate that using more than 100k samples for both FlowGen and FlowDiffusion is crucial to avoid overfitting in the case of human faces. Results are shown in Fig.\ref{fig:supdataset_ablation_flowgen} and Fig.\ref{fig:supdataset_ablation_flowdiffusion}. Please note that although we use L2 reconstruction loss for our final FlowGen model, the ablations above and experiment in Fig.~\ref{fig:flow_normalization} were performed using L1 reconstruction loss. While both L1 and L2 losses exhibit a similar trend, we chose L2 for our final model as it produced slightly more continuous dense flows.

\end{document}